\documentclass{article}

\usepackage{arxiv}

\usepackage{times}

\usepackage{amsmath,amssymb,amsfonts,amsthm}
\usepackage[linesnumbered,ruled,vlined]{algorithm2e} 
\usepackage{algorithmic}
\usepackage{graphicx}
\usepackage{textcomp}
\usepackage{soul}
\usepackage{comment}
\usepackage{bm}
\usepackage{array, makecell}
\usepackage{dsfont}
\usepackage{ulem}
\usepackage{optidef}
\usepackage{booktabs}
\usepackage{multicol}
\usepackage{subfig}
\usepackage[dvipsnames]{xcolor}
\definecolor{cvprblue}{rgb}{0.21,0.49,0.74}
\usepackage[pagebackref,breaklinks,colorlinks,allcolors=cvprblue]{hyperref}
\usepackage{multirow}
\usepackage{tcolorbox}
\usepackage{makecell}
\usepackage{pifont}
\usepackage[numbers,sort&compress]{natbib} 
\newcommand{\cmark}{\ding{51}}%
\newcommand{\xmark}{\ding{55}}%

\newtheorem{Def}{Definition}

\SetCommentSty{mycommfont}

\SetKwInput{KwInput}{Input}                
\SetKwInput{KwOutput}{Output}              
\SetKwRepeat{Do}{do}{while}%

\DeclareMathAlphabet{\mathbbmsl}{U}{bbm}{m}{sl}

\title{Can Large Vision Language Models Read \\Maps like a Human?}

\author{
\bf
Shuo Xing$^1$\thanks{\  Equal Contribution.},\quad
Shuangyu Xie$^2$\footnotemark[1],\quad
Zezhou Sun$^3$\footnotemark[1],\quad
Kaiyuan Chen$^2$,\quad
Yanjia Huang$^1$,\\ 
\bf  Yuping Wang$^4$, \quad
Jiachen Li$^5$,\quad
Dezhen Song$^3$,\quad
Zhengzhong Tu$^1$\thanks{\  Corresponding author.}\\
\\
$^1$Texas A\&M University \quad
$^2$UC Berkeley \quad
$^3$MBZUAI \\ 
$^4$University of Michigan \quad
$^5$UC Riverside\\
\\
\texttt{\{shuoxing, tzz\}@tamu.edu}
}

\renewcommand{\headeright}{Preprint}

\renewcommand{\shorttitle}{Can Large Vision Language Models Read Maps like a Human?}

\hypersetup{
pdftitle={Can Large Vision Language Models Read Maps like a Human?},
}

\begin{document}
\maketitle

\begin{center}
    \includegraphics[width=0.95\textwidth]{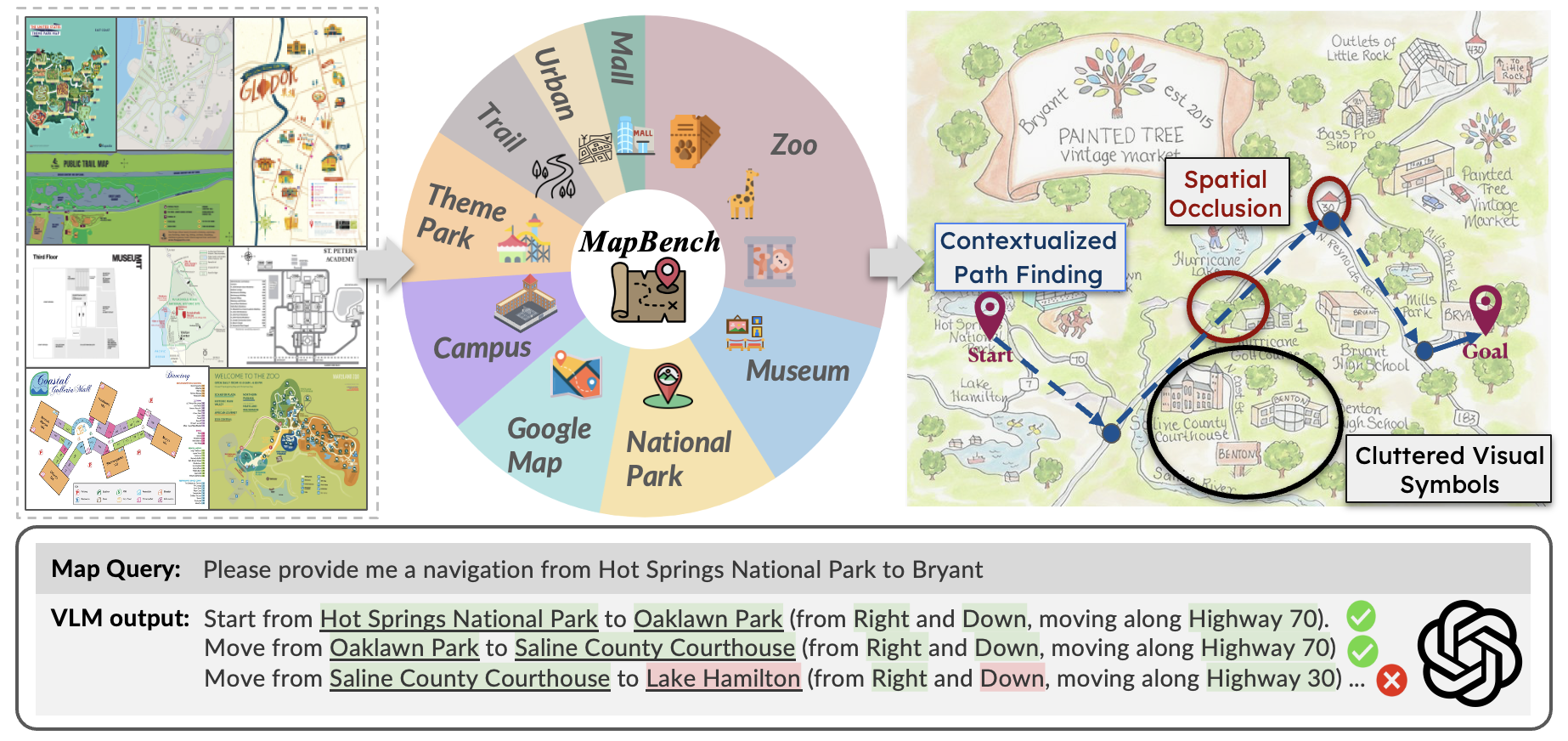}\vspace{-2mm}
    \captionof{figure}{
    \textbf{MapBench} is a dataset of over 1600 map space path-finding problems from 100 diverse map images. 
    MapBench evaluates language-based navigation instructions generated by Large Vision-Language Models (LVLMs) with map images with cluttered and potentially occluded visual symbols. 
    }
    \label{fig:teaser}
\end{center}

\vspace{2em}

\begin{abstract}
In this paper, we introduce \textbf{MapBench}—the first dataset specifically designed for human-readable, pixel-based map-based outdoor navigation, curated from complex path finding scenarios. 
\textbf{MapBench} comprises over 1600 pixel space map path finding problems from 100 diverse maps.
In MapBench, LVLMs generate language-based navigation instructions given a map image and a query with beginning and end landmarks. 
For each map, \textbf{MapBench} provides Map Space Scene Graph (MSSG) as an indexing data structure to convert between natural language and evaluate LVLM-generated results.
We demonstrate that MapBench significantly challenges state-of-the-art LVLMs both zero-shot prompting and a Chain-of-Thought (CoT) augmented reasoning framework that decomposes map navigation into sequential cognitive processes. 
Our evaluation of both open-source and closed-source LVLMs underscores the substantial difficulty posed by MapBench, revealing critical limitations in their spatial reasoning and structured decision-making capabilities. We release all the code and dataset in \url{https://github.com/taco-group/MapBench}.

\end{abstract}

\section{Introduction}

\textit{``Not all those who wander are lost.''}
\hfill -- J. R. R. Tolkien

Maps have guided travellers since ancient times. From traditional nautical charts to modern digital navigation systems, our capacity to transform two-dimensional pixel representations into actionable navigation instructions remains a distinctly human skill.
The recent emergence of Large Vision-Language Models (LVLMs)~\cite{li2022blip, li2023blip2, liu2024llava,llavanext,llama3.2, Qwen-VL, Qwen2VL, lu2024deepseek,wu2024deepseek}  
reshapes the boundaries of human and machines in visual perception~\cite{huang2023language,tu2021ugc,zheng2024video} and language understanding~\cite{radford2018improving,yang2019xlnet} across various domains, including biomedical imaging~\citep{moor2023med,li2024llava-med,zhang2023biomedgpt}, autonomous vehicles~\citep{shao2024lmdrive,xing2024autotrust,tian2024drivevlm,sima2023drivelm,xing2025openemma}, robotics~\citep{rana2023sayplan, goldberg2024blox,kim2024openvla,zhang2024grape}, and sciences~\citep{}
In this paper, we explore whether LVLMs can replicate the human cognitive ability of \textit{map-space path finding}: given a human-readable map image with landmarks and pathways, and a query with beginning and end landmarks, LVLM generates language-based navigation instructions. 
Effective map-space path finding by LVLMs necessitates three fundamental capabilities:
(a) perception that recognizes visual symbols such as colors, texts, areas, and icons in artistic or stylized forms;
(b) spatial understanding that contextualizes the symbols to physical environment, addressing orientations, viewpoint, occlusion handling, and scaling;
(c) planning that routes between endpoints through landmarks and intersections.
While research has advanced each individual component independently~\cite{liu2024ocrbench, chen2024spatialvlm, wang2024picture, zhang2025mapnav, goetting2024end}, the integration of these capabilities must simultaneously interpret visual symbolic representations and extract spatial relationships to enable coherent path finding.
Existing LVLMs exhibit limitations in cross-modal misalignment~\citep{rohrbach2018object,bai2024hallucination,realign,xing2024autotrust} and understanding complex geometric relationships within visual inputs~\citep{gao2023gllava}. 
As a result, it is critical to investigate LVLM capabilities specifically as a \textit{unified} context of map-space path finding, effectively bridging visual perception, spatial reasoning and route planning.

We study the map-space path finding task for VLM with Map Space Scene Graph (MSSG), a structured representation that captures visual symbolic-spatial-topological relationship of human-readable maps. 
Based on MSSG, we introduce MapBench, the first comprehensive dataset specifically designed to evaluate LVLMs on human-readable maps. MapBench encompasses over 1,600 visual path planning queries derived from 100 distinct maps, categorized across 4 visual style elements (landmarks, traversable areas, projection methods, and annotation formats) and 9 scenarios (urban, university, theme park, etc.). Each map includes manually annotated MSSG along with associated queries of beginning and end landmarks. 
We evaluate multiple state-of-the-art LVLMs on MapBench, including Llama-3.2-11B-Vision-Instruct~\citep{llama3.2}, Qwen-2-VL-7B~\citep{Qwen2VL}, GPT-4o~\citep{gpt4o}, and GPT-4o mini~\citep{gpt4omini} under both zero-shot prompting and chain-of-thought reasoning protocols. 

To further investigate the performance in map-space path finding, we propose a Chain-of-Thought (CoT) augmented reasoning framework that aligns with the proposed MSSG by decomposing map navigation into sequential cognitive processes, including (1) localization of start and destination landmarks; (2) description of surrounding context; (3) path connection through identifying intermediate landmarks to enable LVLMs to perform explicit spatial reasoning. 
Our results indicate that MapBench poses significant challenges for state-of-the-art LVLMs. Additionally, closed-ended LVLMs outperform their open-sourced LVLMs. The proposed CoT reasoning framework generally delivers better performance compared to zero-shot prompting, though it occasionally introduces redundant information.

In summary, our contribution is as follows:

\begin{itemize}
    \item Map Space Scene Graph (MSSG), a visual symbolic-spatial-topological representation of maps, 
    \item MSSG-based LVLMs evaluation metrics and conversion algorithms between MSSG and natural language,
    \item MapBench, a human readable map benchmark with 1600 human-annotated map queries 
    \item A CoT-augmented reasoning framework that decomposes map navigation into sequential prompting,
    \item Evaluation results of LVLMs on MapBench with both zero-shot and Chain-of-Thought augmented prompting. 
\end{itemize}

\section{Related Works}

\paragraph{Vision Language Model Reasoning and Planning}
Chain-of-Thought (CoT), whereby the simple inclusion of accurate deduction steps for few-shot examples within the original prompt empowers LLMs to achieve substantial performance improvements in reasoning tasks~\citep{wei2022cot}, pioneers the revolution reasoning and planning tasks capabilities of large language models and vision language models by prompting based strategies~\citep{pan2023plum, pryzant2023automatic, yao2023tot, xu2024reprompt, Ye2024, Zhang2023Feb, rose2023visual}. Moreover, integrating external tools, knowledge bases, and iterative self-reflection enhances the robustness and adaptability of VLMs, driving them toward more sophisticated, human-like visual intelligence in complex real-world tasks~\citep{xing2025openemma, hwang2024emma, sima2024drivelm, rana2023sayplan, kim2024openvla, xi2025rise}.

\paragraph{Visual Language Navigation} Visual Language Navigation (VLN) tasks require agents to interpret natural language instructions to navigate through photorealistic or real-world environments. Some previous work has mad significant progress in field. For example, leveraging large language models (LLMs) like NavGPT enhances instruction understanding and reasoning capabilities, enabling zero-shot sequential action predictions by decomposing instructions into sub-goals and incorporating commonsense knowledge \cite{zhou2023navgptexplicitreasoningvisionandlanguage} \cite{goetting2024endtoendnavigationvisionlanguage} \cite{buoso2024select2plantrainingfreeiclbasedplanning}. Additionally, frameworks such as DAVIS utilize visual consistency to anticipate discrepancies in unseen environments by combining semi-supervised learning with reinforcement learning and integrating online visual-language mapping techniques, allowing agents to parse natural language instructions into executable plans aligned with real-time visual inputs, thereby facilitating better adaptation during performance \cite{lu2022anticipatingunseendiscrepancyvision} \cite{xu2023visionlanguagenavigationreal} \cite{gao2024aerialvisionandlanguagenavigationsemantictopometric} \cite{hong2024navigatinginstructionsvisionandlanguagenavigation}. Despite these advancements, even with the use of frontier-based exploration methods in language-driven zero-shot object navigation \cite{wen2024zeroshotobjectnavigationvisionlanguage} \cite{guan2024loczsonlanguagedrivenobjectcentriczeroshot}, existing approaches still heavily rely on detailed visual inputs, including depth and semantic information, and struggle to generalize to unseen environments due to discrepancies between training and testing conditions. Moreover, they seldom utilize high-level, human-readable maps that humans commonly rely on for navigation. 

\paragraph{VQA reasoning and planning benchmarks} Visual Question Answering (VQA) has been a widely studied task in the field of computer vision and natural language processing, with the aim of enabling models to answer questions about images. Prior VQA benchmark extends from CV dataset focusing on evaluating simple visual-language correlation task, such as image classification (VQA \cite{antol2015vqa}, MSCOCO~\cite{chen2015microsoftcoco}).  The general VQA benchmark covers domain specific tasks such as (MM-Vet~\citep{yu2023mm}, MMBench~\cite{liu2024mmbenchmultimodalmodelallaround}, MMMU~\cite{yue2023mmmu}, MMMU-Pro~\citep{yue2024mmmu}). Reasoning about spatial relationships using images with simple context is exemplified in works such as MathVista~\cite{lu2023mathvista}, ChartQA~\cite{masry2022chartqa}, and ScienceQA~\citep{lu2022scienqa}. We focus on visually complicated map reading task that has multimodality unstructured information. 

\begin{table}[htbp]
  \footnotesize
  \setlength{\tabcolsep}{4pt}

  \begin{center}
    \begin{tabular}{lcccccccccc}
      \toprule

      Dataset
      & Perception
      & \makecell[c]{Text\\Recognition}
      & \makecell[c]{Spatial\\Reasoning}
      & \makecell[c]{Long\\Horizon}
      \\

      \midrule

      ScienceQA~\citep{lu2022scienqa}
      & \cmark
      & \xmark
      & \xmark
      & \xmark
      \\

      TextVQA~\citep{singh2019textvqa}
      & \cmark
      & \cmark
      & \xmark
      & \xmark
      \\

      VizWiz~\citep{gurari2018vizwiz}
      & \cmark
      & \xmark
      & \xmark
      & \xmark
      \\

      MM-Vet~\citep{yu2023mm}
      & \cmark
      & \cmark
      & \xmark
      & \xmark
      \\

      MMMU~\citep{yue2023mmmu}
      & \cmark
      & \cmark
      & \xmark
      & \xmark
      \\

      MMMU-Pro~\citep{yue2024mmmu}
      & \cmark
      & \cmark
      & \xmark
      & \xmark
      \\

      \textbf{MapBench}
      & \cmark
      & \cmark
      & \cmark
      & \cmark
      \\
      
      \bottomrule
    \end{tabular}
  \end{center}
  \caption{The comparison between MapBench and existing LVLM VQA benchmarks.}
  \label{tab:comp}
\end{table}

\section{Map Space Scene Graph}
We present a map space scene graph (MSSG) as a indexing data structure for the human readable map. We provide functions that enabling conversion from language instruction to visual map and visual map to language. 

\subsection{MSSG Definition} \label{Sec:MSSG}
For each map image $I_{i} \in \mathcal{D}_m$, we manually construct an outdoor scene graph $\mathcal{G}_i = (V, E)$ that captures \textit{semantic-spatial-topology} relationships between different locations. In the graph $\mathcal{G}_i$, vertices are the locations and key intersections/road crossings $V = V_l \cup V_c$ and edges $E$ reflect the semantics of connectivity.   

Specifically, we define the landmark node $v_l = (x,y,r,s)\in V_l$ that contains an image pixel coordinate $(x,y)\in I_m$ reflecting the center of the landmark, a radius $r \in \mathbb{R}$ with center coordinate that defines a covering circle for the landmark, and a semantic label $s \in \mathcal{S}$ referring to the name of the location, e.g. $s=\texttt{\small "castle"}$.  The intersection node $v_c = (x,y,s)\in V_c$  that contains an image pixel coordinate $(x,y)\in I_m$ reflecting the center of the intersection region and a semantic label $s = \texttt{\small "intersection"}$. We define a fully connected graph and for every edges, $e = (v_i,v_j, c) \in E$, we define $c$ as a connectivity attribute: 
\begin{align} \nonumber
   c =  \begin{cases}
\texttt{\small "connect"}, ~~&\text{if}~  v_i, v_j \in V_c  \wedge \mathds{1}_{\mbox{\tiny road}}(v_i,v_j), \\ 
\texttt{\small "adjacent"}, ~~&\text{if}~ { (v_i\in V_l \vee v_j \in V_l)} \wedge \mathds{1}_{\mbox{\tiny adj}}(v_i,v_j), \\
\texttt{\small "observable"}, ~~&\text{if}~ { (v_i\in V_l \vee v_j \in V_l)} \wedge \mathds{1}_{\mbox{\tiny obs}}(v_i,v_j), \\
\texttt{\small "unrelated"},
~~~&\text{Otherwise.}
\end{cases}
\end{align}
where $\mathds{1}_{\mbox{\tiny road}}(v_i,v_j)$ is the indicator function of road existence between $v_i$ and $v_j$, $\mathds{1}_{\mbox{\tiny adj}}(v_i,v_j)$ is the indicator function of adjacency between $v_i$ and $v_j$.

Note that the semantic label is for convenience of interpreting the scene graph into a language description \cite{krishna2016visualgenomeconnectinglanguage}. 
An example of a scene graph is shown in Fig.~\ref{fig:scene-graph}:

\begin{figure}
    \centering
    \includegraphics[width=0.5\linewidth]{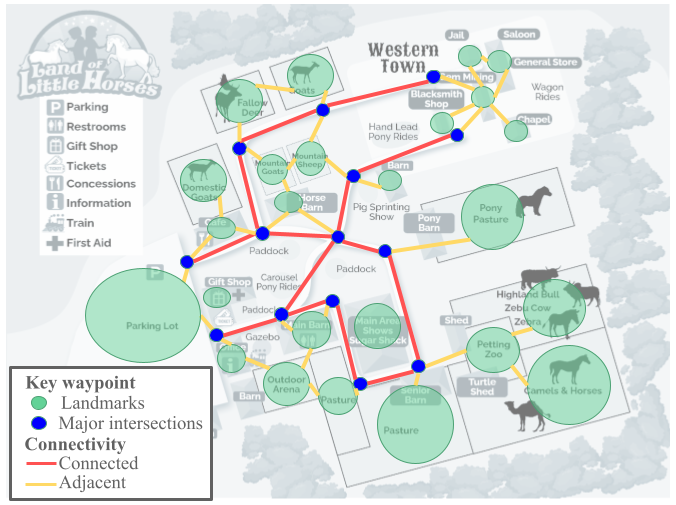}
    \caption{Map Space Scene Graph for a human-readable map.}
    \label{fig:scene-graph}
\end{figure}

\begin{Def}\textbf{MSSG Connectivity}
\begin{itemize}
    \item Every landmark node is at least adjacent to one intersection node. $deg(v_l) \ge 1$
    \item Every intersection node is at least connected to one intersection node. $deg(v_c) \ge 1$
    \item Except for end intersection node, every other intersection node should connect to two other intersection node. 
\end{itemize}
\end{Def}

\noindent The connectivity conditions ensure there is at least one path that exists in the graph for any given start node and end node pair.


\subsection{MSSG-Language Conversion}

\subsubsection{MSSG to Language}
Given a path on MSSG, we convert it to a language description. The algorithm (Alg.~\ref{alg:MSSG}) iteratively converts the edges to language directives from the start node to the end node. In the conversion, there are three different types of language template functions that handle the road walking description (\textbf{describeWalk()}), the move description from a landmark to a road intersection (\textbf{describeMove()}), and a surrounding description function (\textbf{describeSurrounding()}). After the conversion, the language description for each edge is concatenated together as a navigation narrative.

\SetKwComment{Comment}{/* }{ */}

\begin{algorithm}
\caption{MSSG Path Parser}\label{alg:MSSG} 

\small

\KwInput{$G_m, v_s, v_e$, path}
\KwOutput{language-path}

\DontPrintSemicolon


language-path := `'

\Comment{\hfill language conversion \hfill}

$e$ = \textbf{findStartEdge}($v_s$, path)

\While{True} 
{
\Comment{parsing each edge}
direction = \textbf{getDirection}($e.v_i, e.v_j$) 

\If {$e.c == \texttt{\small 'connected'}$ }
{

\Comment{Walk along the road}

 walk = \textbf{describeRoad}(direction,$v_i,v_j$)
 

}
\Else{

\Comment{Move between landmark and  node}

walk = \textbf{describeMove}(direction,$v_i,v_j$)

}
surrounding = \textbf{describeSurrounding}($G_m$,$e$)

language-path += (walk + surrounding)

\If{$e.v_j == v_e$}{
language-path += arrival-note 

\break}

$e$ = \textbf{findNextEdge}($e$, path)

}

\end{algorithm}

\subsubsection{Language to MSSG Path}

To convert a natural language navigation instruction into a structured MSSG path, we employ a systematic parsing approach. Given an annotated map image  $I_i$  and its corresponding MSSG $G_i$ , we use Alg.~\ref{alg:language-parser} to extract meaningful spatial relationships and transform them into a structured path representation. The process begins with \textbf{getLocations(),} which identifies the locations mentioned in the textual instruction and maps them to corresponding nodes in the MSSG. Next, \textbf{split2edges()} establishes connections between these nodes by analyzing spatial relationships described in the instruction, effectively segmenting the parsed locations into a sequence of connected edges. Finally, \textbf{getPath()} constructs a navigable MSSG path by linking these edges into a coherent route that aligns with the original language input. This conversion enables LVLMs to reason about navigation in a structured, graph-based manner, improving their ability to process and interpret complex map-based pathfinding tasks. By structuring the navigation process into nodes (landmarks and intersections) and edges (connections between locations), this algorithm facilitates a machine-readable, scalable, and spatially accurate representation of human-readable navigation instructions.

\begin{algorithm}
\caption{Language Path Parser}\label{alg:language-parser} 
\small

\KwInput{language-path, $\mathcal{G}_i$}
\KwOutput{path}

loc := \textbf{getLocations}(language-path, $\mathcal{G}_i$) \tcp*{Nodes}

roads := \textbf{split2edges}(loc, language-path, $\mathcal{G}_i$) \tcp*{Edges}

path := \textbf{getPath}(loc, roads)

\small

\DontPrintSemicolon

\end{algorithm}

\section{\textsc{MapBench}}
In this section, we present \textsc{MapBench}, a diverse and comprehensive benchmark designed to evaluate the performance of LVLMs in map-space path finding tasks across a wide range of artistic illustrated human-readable maps $\mathcal{D} = \{I\}$. Each query consists of an image of a map, a specified starting point, and a destination. The goal is to assess the model’s ability to perceive and interpret visual input, understand map structures, reason about spatial relationships, and generate an accurate and efficient path between the given points.

\subsection{Data Collection and Curation}
As a first-of-its-kind benchmark, \textsc{MapBench} comprises a curated collection of 100 high-quality map images spanning nine distinct types based on usage scenarios: Zoo, Museum, National Park, Campus, Google Maps, Theme Park, Trail, Urban, and Mall. For each map image, we utilize LabelMe \cite{russell2008labelme} to manually construct the ground truth MSSG introduced in Section \ref{Sec:MSSG}. This process enables the extraction of key spatial and semantic elements, including landmarks, pathways, intersections, and region boundaries, providing a structured representation of the map's information. To ensure a rigorous and reliable evaluation, we use GPT-4o as a reference standard, generating 20 randomly selected start-destination pairs per map, each repeated three times. We then filter out queries that do not receive valid responses and use the remaining ones to construct the final benchmark queries for each image.  

\begin{figure*}
    \includegraphics[width=0.99\textwidth]{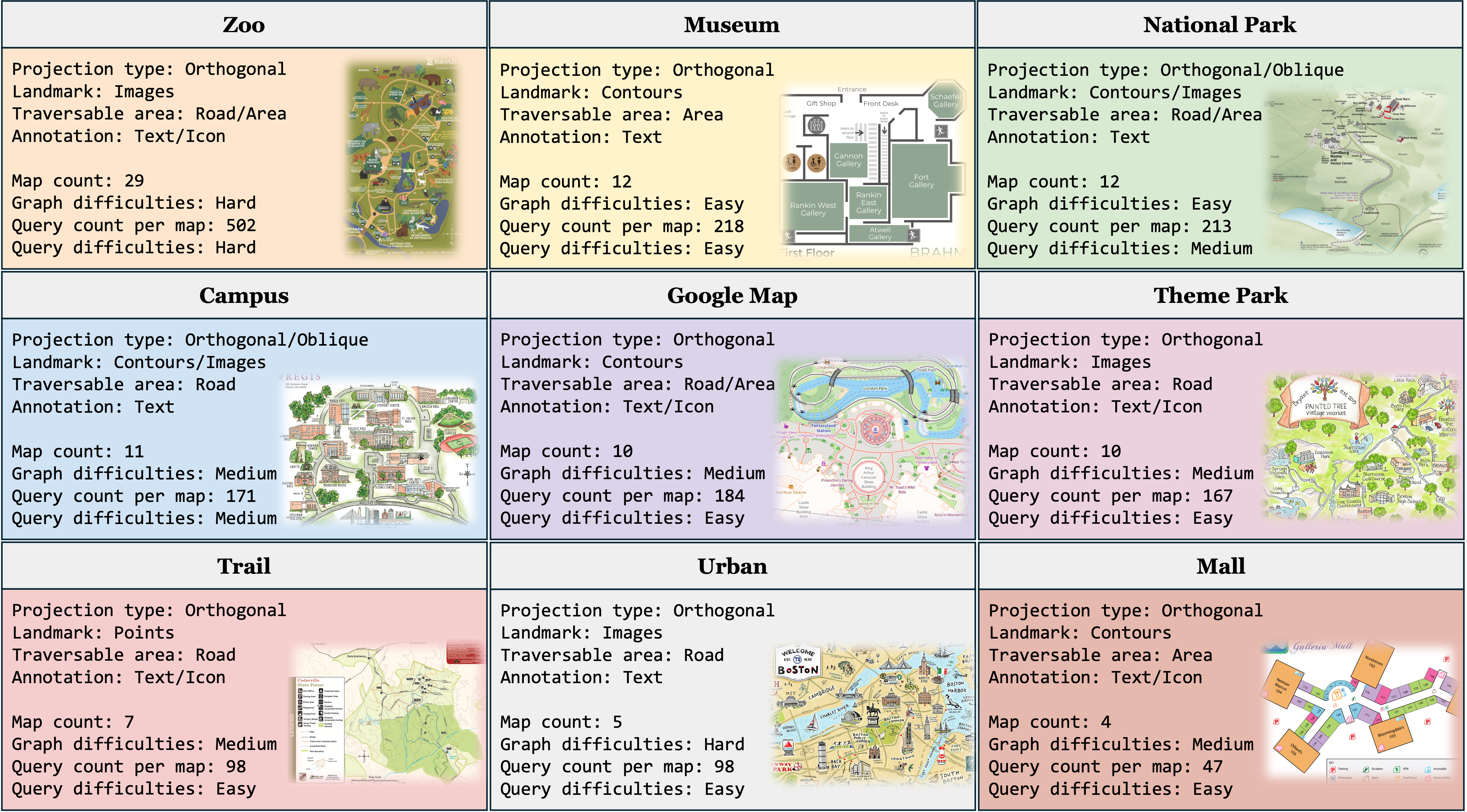}\vspace{-2mm}
    \captionof{figure}{
    Sampled MapBench examples from each scenario. 
    Segmenting the map and navigating based on the query require expert-level spatial reasoning and understanding.
    }
    \label{fig:detail}
\end{figure*}

In total, we compile a final dataset of 1649 instances, with detailed statistics presented in Appendix~\ref{app:map-stats}. To further characterize the dataset, we categorize maps according to three fundamental attributes: landmark representation, traversable area type, and map projection, as illustrated in Figure \ref{fig:detail}.
\begin{itemize}
\item \textbf{Landmark Representation:} Landmarks are categorized as points, contours, or images, each influencing how models segment and interpret the map.
\item \textbf{Traversable Areas:} These are classified into roads and areas—roads feature well-defined intersections, while areas do not, which may impact the model’s ability to discern connectivity between landmarks.
\item \textbf{Map Projection:} Maps are categorized as orthographic or oblique projections, as projection type may affect the model’s perception of spatial relationships and distances.
\end{itemize}





\subsection{Metrics}

In this section, we introduce a set of metrics designed to characterize the unique instructions and features of human-readable maps, providing a comprehensive framework for their evaluation. First, to capture image complexity and query difficulty, we define several key indices, including the Elements Index, Meshedness Index, Average Shortest Path Length Index, and Query Difficulty Index, as detailed in Section~\ref{Sec:taskMetric}. These metrics quantify the structural properties of maps and the challenges they pose for LVLM-based pathfinding (details statistics can be found in Table \ref{tab:stats}). Additionally, we establish performance evaluation metrics in Section~\ref{sec:perf-metric}, which assess the accuracy and effectiveness of model predictions. These include Landmark/Roadname Accuracy, which measures the precision of textual map references, Path Quality Score, which evaluates the optimality of generated routes.

\subsubsection{Task Complexity} \label{Sec:taskMetric}
Task complexity is determined by both graph complexity and query difficulty. For graph complexity, we evaluate it across three dimensions: the number of graph elements, the degree of connectivity, and the overall graph scale.

We define Elements Index to quantify the total number of nodes, representing landmarks and intersections, along with the total number of edges in a given graph.

\begin{Def}{Elements Index.}
For a graph $\mathcal{G} = (V, E)$ :
    \begin{equation}
        EI = |V| + |E|
    \end{equation}
\end{Def}


As for connectivity, we define the Meshedness Index to evaluate the number of cycles in a graph relative to the maximum possible number of cycles. A higher Meshedness Index indicates a more interconnected network.

\begin{Def}{Meshedness Index \cite{buhl2004efficiency}.}
For a graph $\mathcal{G} = (V, E)$ :
    \begin{equation}
        MI = \frac{|E| - |V| + 1}{2 * |V| - 5}
    \end{equation}
\end{Def}

The Meshedness Index ranges from 0 to 1, providing a measure of connectivity that is independent of the number of nodes in the graph.

Being solely based on the number of nodes and edges, EI and MI indices remain limited in revealing structural differences between networks of equal size. To address this, we define the Average Shortest Path Length Index as a more comprehensive measure of the internal structural differences within a network.

\begin{table}[h]
    \centering
    \setlength{\tabcolsep}{6.pt}
    \footnotesize
    \begin{tabular}{lrrrl}
        \toprule
        Scenes & EI & MI & ASPLI & Difficulty \\
        \midrule
        Google Map & 71.5 & 0.121 & 4.678 & Medium \\
        Mall & 93.5 & 0.023 & 5.839 & Medium \\
        Museum & 59.5 & 0.041 & 4.124 & Easy \\
        National Park & 38.8 & 0.071 & 5.185 & Easy \\
        Theme Park & 59.8 & 0.152 & 4.870 & Medium \\
        Trail & 59.0 & 0.055 & 5.277 & Medium \\
        Campus & 45.5 & 0.150 & 4.180 & Medium \\
        Urban & 71.8 & 0.095 & 6.230 & Hard \\
        Zoo & 102.5 & 0.143 & 5.663 & Hard \\
        \bottomrule
    \end{tabular}
    \caption{The EI, MI and ASPLI index for each map type.}
    \label{tab:scene_metrics}
\end{table}

\begin{Def}{Average Shortest Path Length Index \cite{rodrigue2020geography}.}
For a graph $\mathcal{G} = (V, E)$ :
    \begin{equation}
        ASPLI = \frac{1}{|V| * (|V| - 1)}\sum_{\substack{i \neq j}} d(v_{i}, v_{j})
    \end{equation}
\end{Def}
Here, $d(v_{i}, v_{j})$ denotes the shortest distance between $v_{i}$ and $v_{j}$, where $v_{i}, v_{j} \in V$ 

Similarly, we define the Query Difficulty Index to considers the difficulty of solving or navigating the query case.

\begin{Def}{Query Difficulty Index.}
For a query in the graph described by $(G, s, t)$ :
    \begin{equation}
        QDI(G, s, t) = \frac{1}{N} \sum_{i=1}^{N} \ell(L_{i})
    \end{equation}
\end{Def}
Here, $L_{i}$ denotes a simple path from source $s$ to target $t$. $N$ is the total number of all simple paths from $s$ to $t$, where simple path \cite{sedgewick2001algorithms} is a path with no repeated nodes. $\ell(L)$ denotes the path length.



\begin{table}[h]
\footnotesize
    \centering
    \begin{tabular}{lrrrrl}
        \toprule
        Scenes & Mean & Min & Max & Variance & Difficulty \\
        \midrule
        Google Map & 11.45 & 4.00 & 21.01 & 29.82 & Easy \\
        Mall & 10.47 & 5.50 & 20.17 & 47.00 & Easy \\
        Museum & 6.93 & 1.71 & 12.77 & 8.65 & Easy \\
        National Park & 12.43 & 2.37 & 33.11 & 96.15 & Medium \\
        Theme Park & 9.62 & 1.44 & 18.73 & 50.29 & Easy \\
        Trail & 9.60 & 2.77 & 22.21 & 55.69 & Easy \\
        Campus & 13.45 & 2.13 & 23.00 & 46.86 & Medium \\
        Urban & 10.07 & 2.76 & 19.99 & 52.91 & Easy \\
        Zoo & 18.71 & 2.04 & 41.43 & 145.04 & Hard \\
        \bottomrule
    \end{tabular}
    \caption{Query Difficulty statistics for each map type.}
    \label{tab:query_difficulty}
\end{table}

\subsubsection{Performance Metrics}\label{sec:perf-metric}

Given a language path, we convert it to MSSG path and compare it with the MSSG graph path. We can evaluate the path feasibility and quality metrics including connectivity descriptions and format accuracy.

\paragraph{Language Description Accuracy}



We evaluate whether the model correctly identifies and describes language paths. The assessment consists of three levels: path existence check, linguistic coherence check, and format compliance check:

\begin{itemize}
    \item Missing Paths refer to cases where the model either fails to output a route or generates a response indicating that no path is found.
    \item Linguistic Incoherence occurs when the generated route descriptions are discontinuous, such as inconsistencies between the start and end points or interruptions in the middle of a route.
    \item Format Non-Compliance refers to cases where the route descriptions do not adhere to predefined formatting requirements, such as failing to use landmarks for path descriptions or relying on informal phrasing instead.
\end{itemize}
We calculate the proportion of errors for each of these three categories, providing insights into the system’s robustness in generating accurate and comprehensible navigation instructions.

\paragraph{Path Quality Score}

We define the Path Quality Score based on the ratio of the length of the MSSG path to the length of the shortest path.

\begin{Def}{Path Quality Score }
    For a query in the graph described by $(G, s, t)$. 
    \begin{equation} 
        PQS(G, s, t) = \frac{\ell(L_{MSSG})}{\ell(L_{sp})} 
    \end{equation} 
\end{Def}
Here, $L_{sp}$ denote the shortest path from source $s$ to target $t$, and $L_{MSSG}$ denote the MSSG path.

\subsection{Comparison with Existing Benchmarks}
MapBench is the first comprehensive and specialized benchmark focusing on the map-space pathfinding, featuring high-quality human annotations. Unlike the previous VQA benchmarks, which primarily focus on static object recognition and short-form question-answering, MapBench integrates spatial reasoning and long horizon planning, making it uniquely suited for assessing LVLMs' capabilities in real-world navigation tasks. Furthermore, MapBench pioneers the first structurally-grounded human annotation framework for pathfinding tasks — MSSG. It not only provides a precise and effective evaluation of model performance but also serves as a valuable tool for future research and development in map-based reasoning and navigation, bridging the gap between topological understanding and multimodal AI. The detailed comparsion can be found in Tab.~\ref{tab:comp}.

\begin{figure*}[htbp]
  \centering
   \includegraphics[width=.9\linewidth]{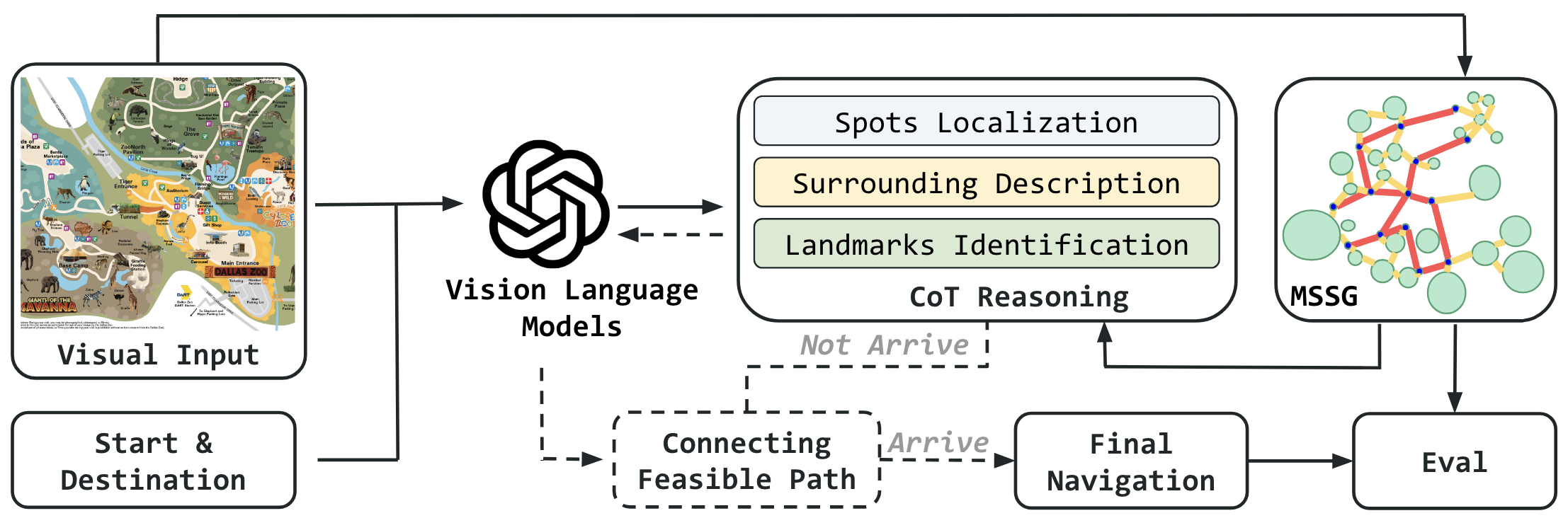}
   \vspace{-2mm}
  \caption{Illustration of CoT extension to VLMs. }
  \label{fig:pipeline}
\end{figure*}

\section{Experiments}

\begin{table*}[h]
\footnotesize
  \begin{center}
  \setlength{\tabcolsep}{5.pt}
    \begin{tabular}{llccccccccc}
      \toprule
        Method
        & Model
        & Google Map
        & Mall
        & Museum
        & National Park
        & Theme Park
        & Trail
        & Campus
        & Urban
        & Zoo
        \\

        \midrule

        \multirow{4}{*}{\textit{Zero-shot}} 
        & Llama-3.2
        & 1.868
        & 3.962
        & 2.652
        & 3.859
        & 3.401
        & 3.121
        & 2.195
        & 2.194
        & 1.989
        \\
        & Qwen2-VL
        & 3.289
        & 2.363
        & 2.817
        & \textbf{2.154}
        & 2.027
        & 2.188
        & 2.459
        & 3.840
        & 1.961
        \\
        & GPT-4o mini
        & 2.755
        & 3.086
        & 3.526
        & 2.367
        & \textbf{1.327}
        & 2.866
        & 4.032
        & 2.084
        & 2.084
        \\
        & GPT-4o
        & \textbf{1.670}
        & 2.372
        & \textbf{1.706}
        & 2.372
        & 1.436
        & 2.766
        & 2.133
        & 2.755
        & \textbf{1.886}
        \\

        \midrule

        \multirow{4}{*}{\textit{CoT}} 
        & Llama-3.2
        & 2.095
        & 3.366
        & 3.196
        & 2.859
        & 2.995
        & 2.121
        & 3.328
        & 3.153
        & 2.365
        \\
        & Qwen2-VL
        & 2.532
        & 2.087
        & 1.879
        & 2.308
        & 1.577
        & 2.088
        & 2.203
        & 2.951
        & 2.213
        \\
        & GPT-4o mini
        & 2.277
        & 4.182
        & 3.215
        & 2.222
        & 1.520
        & \textbf{1.213}
        & 4.037
        & \textbf{1.855}
        & 2.067
        \\
        & GPT-4o
        & 1.820
        & \textbf{1.825}
        & 2.142
        & 2.766
        & 1.874
        & 1.884
        & \textbf{1.865}
        & 2.561
        & 2.040
        \\ 
      \bottomrule
    \end{tabular}
  \end{center}
  \caption{The averaged score of the SOTA LVLMs on MapBench under zero-shot prompting and CoT reasoning.}
  \label{tab:ade}
\end{table*}

\begin{algorithm}
\caption{Overview of CoT reasoning}\label{alg:goat-general} 

\small

\KwInput{$I, L_s, L_e, \mathcal{P}$, VLM}
\KwOutput{readable-navigation}

\DontPrintSemicolon


readable-navigation := `'

\Comment{\hfill language conversion \hfill}

location-information = \textbf{Localization}(VLM, $I, L_s, L_e, \mathcal{P}$)

\textbf{concat}($\mathcal{P}$, location-information)

surrounding = \textbf{describeSurrounding}(VLM, $I, L_s, \mathcal{P}$)

\textbf{concat}($\mathcal{P}$, surrounding)

readable-navigation := `'

\While{True}{
navigation = \textbf{connectPath}(VLM, $I, L_s, L_e, \mathcal{P}$)

\textbf{concat}(readable-navigation, navigation)

\If{reach $L_e$}{
\textbf{break}
}
}
\Comment{Connecting each intermediate landmark and major intersection into a feasible path}

\textbf{concat}($\mathcal{P}$, detailed-navigation)

readable-navigation = \textbf{Summarize}(VLM, $I, \mathcal{P}$)

\end{algorithm}

In this section, we conduct a comprehensive empirical analysis of state-of-the-art vision-language models (VLMs) on our proposed MapBench, assessing both open-source and proprietary LVLMs (detailed information can be found in the appendix).

\paragraph{Setup}
For all evaluation experiments, we take the input maps as the visual input of the LVLMs, along with the textual instructions. The evaluated models spans both open-sourced and proprietary LVLMs:
\textbf{Llama-3.2-11B-Vision-Instruct}~\citep{llama3.2}, \textbf{Qwen-2-VL-7B}~\citep{Qwen2VL},\textbf{ GPT-4o}~\citep{gpt4o}, and \textbf{GPT-4o mini}~\citep{gpt4omini}). To ensure comprehensive analysis, we assess model performance under two distinct settings:
\begin{itemize}
    \item \textbf{Zero-Shot Prompting:} The models are prompted to generate accurate and relevant responses based solely on the provided visual inputs (maps) and textual instructions (presented as the following), without any prior specific training on similar tasks.

    \item \textbf{CoT Reasoning:} To further investigate model performance on MapBench, we introduce a CoT reasoning framework that aligns seamlessly with our proposed structural data, aiming to simulate a more deliberative cognitive process. The details of our curated CoT reasoning framework for generating navigation instructions from map inputs are presented in Figure \ref{fig:pipeline} and Algorithm \ref{alg:goat-general}. Generally, the framework is built upon four fundamental, interdependent components as follows:

    \begin{itemize}
        \item \textbf{\textit{Localization}}: Prompt the LVLMs to identify key landmarks and their spatial relationships within the map, associating them with corresponding coordinates. This process allows LVLMs to generate a simplified MSSG, using landmark locations to understand connections and structures within the map.

        \item \textit{\textbf{DescribeSurrounding}} The backbone LVLM is then prompted to generate detailed information about the surroundings of the starting landmark, including their spatial relationships and connectivity, ensuring a comprehensive understanding of the environment. 
 
        \item \textit{\textbf{ConnectPath}} After obtaining the topology and structural information from the constructed simplified MSSG in natural language, the model is tasked with extracting key clues and spatial relationships essential for generating an optimal navigation path (shortest path) between the given start and destination points. This process is completed by iteratively verifying whether the destination has been reached or if further adjustments are required.

        \item \textit{\textbf{Summarize}} The complete navigation route is generated as a human-readable output, which provides a concise yet comprehensive navigation summary. This summary distills the step-by-step route details, ensuring clarity and interpretability for users.
    \end{itemize}

\end{itemize}

Throughout the execution of Algorithm \ref{alg:goat-general}, the textual instruction $\mathcal{P}$ is progressively augmented in alignment with the sequential order of these four components, forming a coherent reasoning chain tailored for the navigation planning task.

\paragraph{Results} The detailed performance of LVLMs on MapBench is shown in Table 1, and our key findings are summarized as follows:

\noindent \textit{\textbf{Challenging Evaluation.}} As shown in Table \ref{tab:ade}, the results indicate the relative quality of the navigation routes generated by the model compared to the ground truth shortest path. Specifically, the evaluation highlights how closely the model-generated routes align with optimal paths in terms of efficiency, providing insights into the model's navigation performance. Notably, the results indicate that the performance of LVLMs falls significantly short of the theoretically optimal navigation, exposing key limitations in multimodal information understanding, spatial reasoning, and decision-making under complex long-horizon planning. The instances of output responses can be found in Appendix \ref{app:smaple}.

\noindent \textbf{\textit{Open-source Models versus Closed-source Models.}} The open-sourced LVLMs generally outperform the closed-source LVLMs, demonstrating superior adaptability and robustness in map-space pathfinding tasks. Among all evaluated models, GPT-4o achieves the highest performance as expected. Furthermore, a huge proportion of queries fail to produce a valid response on navigation when using open-sourced LVLMs, highlighting the challenges these models face in interpreting complex map structures, reasoning over spatial relationships, and providing coherent navigation instructions.

\noindent \textit{\textbf{Performance across Different Prompting Strategy.}} Generally, LVLMs exhibit superior performance on MapBench when using Chain-of-Thought (CoT) prompting compared to zero-shot prompting. CoT prompting enhances spatial reasoning and step-by-step path planning, allowing models to break down complex navigation tasks into interpretable intermediate steps. However, in pathfinding tasks, LVLMs with CoT reasoning often generate redundant information about landmarks and intersections, even when these locations are far from the intended route. Instead of focusing on the optimal and necessary path elements, CoT prompting sometimes overemphasizes peripheral locations, which may distract from efficient route planning. This tendency to generate excessive contextual details is a primary reason why the CoT results in Table \ref{tab:ade} show inconsistencies or suboptimal performance, despite the overall improvements in logical reasoning and interoperability (instance can be found in Appendix \ref{app:smaple}).

\section{Conclusion}\label{sec:con}

In this paper, we introduce MapBench, the first-of-its-kind benchmark designed to evaluate LVLMs in human-readable map-space pathfinding tasks. MapBench comprises 100 high-quality map images spanning nine distinct types based on real-world usage scenarios, including Zoos, Museums, National Parks, Campuses, Google Maps, Theme Parks, Trails, Urban Areas, and Malls, along with 1,649 diverse queries. It establishes a new standard for assessing LVLMs' capabilities in perception, text recognition, spatial reasoning, and long-horizon planning. Additionally, we present the MSSG, a structured indexing data representation that encodes landmarks, paths, and spatial relationships within human-readable maps. To support robust evaluation, we propose a suite of task complexity and performance metrics, designed to assess both the difficulty of maps and queries and the quality of model-generated responses.

\section{Limitations}\label{sec:lim}
As the first benchmark designed to evaluate LVLMs in map-space pathfinding tasks, MapBench has certain limitations that highlight potential areas for future improvement. First, we integrate only 100 high-quality maps with human annotations of Map Space Scene Graph (MSSG). While this dataset covers a diverse range of real-world scenarios, its scale remains limited, and expanding it to include more maps could further enhance its robustness. Additionally, due to the expert-level challenges in MapBench requiring advanced perception, text recognition, spatial reasoning, and long-horizon planning capabilities that are still evolving in current LVLMs, only a small subset of state-of-the-art LVLMs could produce valid responses suitable for evaluation, underscoring the gap between current LVLMs and human-level navigation abilities.

\bibliographystyle{unsrtnat}
\bibliography{main}  






\clearpage
\appendix
{\centering
\Large \vspace{0.5em} \bf Supplementary Material \\
\vspace{1.0em}
}

\section{Map Visual Styles}
\label{sec: Visual Styles}


Each human readable map can be viewed as a combination of content and visual style. For maps presenting identical content, variations in the visual styles of landmarks and traversable areas may hinder a robot’s ability to interpret the map accurately.

From the perspective of the map creation process, we divide it into four phases:
\begin{itemize}
\item \textbf{Sketching:} 
This phase focuses on identifying the outline and location of the landmark. Contours contribute to size and geometric agency, while location allows the reader to clarify spatial relationships with other landmarks.

\item \textbf{Coloring:} 
Color is applied to both landmarks and the ground. For landmarks, color serves as a distinguishing feature between same type of landmarks. For the ground, different colors represent different areas, helping readers identify the area’s characteristics (e.g., river, green space, dining area) and whether it is traversable.

\begin{figure}[ht]
  \centering
   \includegraphics[scale=0.6]{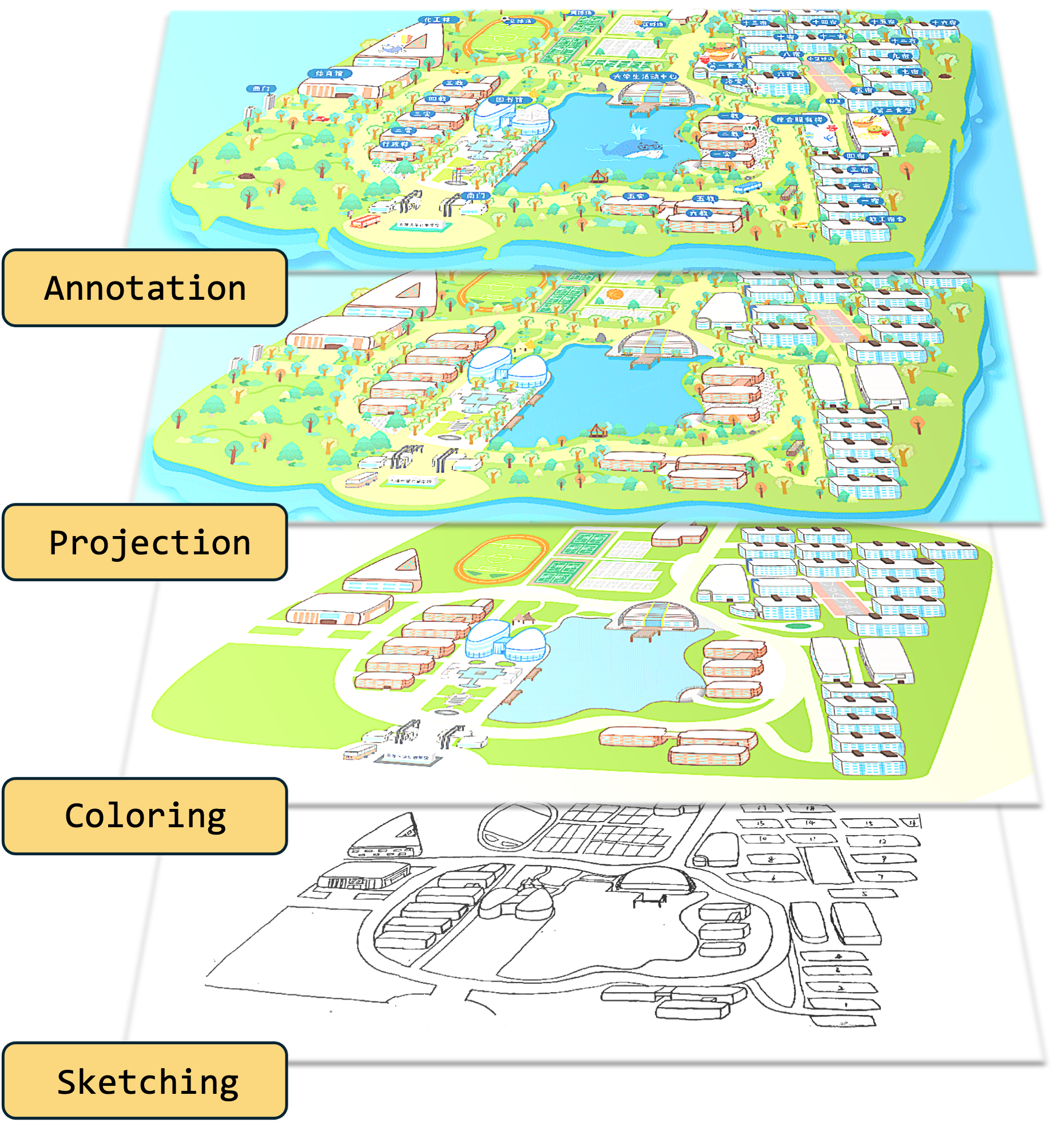}
  \caption{Illustrating the four phases of map creation.}
  \label{fig:painting_steps}
\end{figure}

\item \textbf{Projection:} Projection determines how the three-dimensional world is represented on a two-dimensional map. Different projection methods emphasize various aspects of spatial relationships. Common approaches include orthographic projection, which preserves the shape and proportional relationships of objects, and oblique projection, which provides a certain degree of perspective effect, enhancing depth perception and spatial awareness, making the map more intuitive and readable.

\item \textbf{Annotation:} Symbols and text provide additional information about landmarks, including their names and numbering, and help differentiate same type of landmarks and roads.
\end{itemize}

In summary, we categorized the visual styles of human-readable maps based on how they differentiate landmarks, traversable areas, projection methods, and annotation details. These visual styles influence how planned routes are described in natural language. They are independent of each other and can be combined in various ways within a readable map. However, to ensure consistency, we assume a uniform visual style throughout the map.



\section{Difficulty Classification}

In this section, we explain how we classify graph difficulty and query difficulty into three levels: Easy, Medium, and Hard.

\subsection{Graph Difficulty Classification}

As defined in Section~\ref{Sec:taskMetric}, there are three indexes for evaluating graph difficulty: EI, MI and ASPLI. First, we normalize the three indices by:
\begin{equation}
    \mathcal{X}_{norm} = \frac{\mathcal{X} - \mathcal{X}_{min}}{\mathcal{X}_{max} - \mathcal{X}_{min}}
\end{equation}
Here, $\mathcal{X}$ stands for the set of EI, MI or ASPLI Index.
$\mathcal{X}_{min}$ and $\mathcal{X}_{max}$ are the minimum and maximum values of the indexes, respectively.

We assign the same weight to these three indexes, so the graph difficulty $\mathcal{D}$ is calculate as:
\begin{equation}
    \mathcal{D} = \frac{\mathrm{EI}_{norm} + \mathrm{MI}_{norm} + \mathrm{ASPLI}_{norm}}{3}
\end{equation}

For the average graph difficulty $d_{avg}$ in each scenario , we consider it to be easy for scenarios with $d_{avg} < 0.33$, moderate for scenarios with $0.33 \leq d_{avg} \leq 0.66$, and difficult for scenarios with $d_{avg} > 0.66$.

\subsection{Query Difficulty Classification}

We similarly start by normalizing the query difficulty in each scenario. However, when calculating the query difficulty $q$ for each scenario, we consider both the mean and variance of the query difficulty:
\begin{equation}
    q = \frac{\mathrm{Mean}_{norm} + \mathrm{Variance}_{norm}}{2}
\end{equation}

For the query difficulty $q$ in each scenario , we similarly consider it to be easy for scenarios with $q < 0.33$, moderate for scenarios with $0.33 \leq q \leq 0.66$, and difficult for scenarios with $q > 0.66$.


\section{Map Query Statistics} \label{app:map-stats}
We provides a detailed breakdown of the MapBench dataset, summarizing the distribution of maps and queries across various real-world navigation scenarios. The dataset consists of 100 map images, with 93\% using orthographic projection and 7\% using oblique projection, ensuring diverse spatial representations. It includes a total of 1,649 queries, categorized into nine different map types: Zoo (30.44\%), Museum (13.22\%), National Park (12.92\%), Google Maps (11.16\%), Campus (10.37\%), Theme Park (10.13\%), Trail (5.94\%), Urban (5.94\%), and Mall (2.85\%). The dataset also classifies queries by difficulty levels, with 46.27\% labeled as easy, 23.29\% as medium, and 30.44\% as hard, reflecting the varying complexities of pathfinding tasks. Furthermore, the difficulty of the graph is classified as easy (42\%), medium (24\%), and hard (34\%), based on the structural complexity of the maps. These statistics highlight the diversity and challenge of MapBench, making it a robust benchmark to evaluate the capabilities of LVLMs in human-readable map-based pathfinding.

    \begin{table}[htbp]
  \begin{center}
    \begin{tabular}{lrr}
      \toprule

      Statistics
      & Number
      & Percentage
      \\
      \midrule
      Total Map images
      & 100
      & --
      \\
      \quad * Orthographic Projection&
      93
      &
      93\%
      \\
      \quad * Oblique Projection&
      7
      &
      7\%
      \\
      \midrule
      Total Queries
      & 1649 
      & --
      \\
      \quad * Zoo 
      & 502
      & 30.44 \%
      \\
      \quad * Museum 
      & 218 
      & 13.22 \%
      \\
      \quad * National Park 
      & 213 
      & 12.92 \%
      \\
      \quad * Google Maps 
      & 184 
      & 11.16 \%
      \\
      \quad * Campus 
      & 171 
      & 10.37 \%
      \\
      \quad * Theme Park 
      & 167 
      & 10.13 \%
      \\
      \quad * Trail 
      & 98
      & 5.94 \%
      \\
      \quad * Urban 
      & 98 
      & 5.94 \%
      \\
      \quad * Mall
      & 47 
      & 2.85 \%
      \\
      \midrule
      Query Difficulty: &
      \\
      \quad * Easy
      & 763 
      & 46.27 \%
      \\
      \quad * Medium
      & 384 
      & 23.29 \%
      \\
      \quad * Hard
      & 502 
      & 30.44 \%
      \\
      \midrule
      Graph Difficulty: &
      \\
      \quad * Easy
      & 42 
      & 42 \%
      \\
      \quad * Medium
      & 24 
      & 24 \%
      \\
      \quad * Hard
      & 34 
      & 34 \%
      \\
      \bottomrule
    \end{tabular}
  \end{center}
  \caption{Dataset statistics.}
  \label{tab:stats}
\end{table}

\section{Additional Experiment Results}

Table \ref{tab:failure_rate} presents the failure case analysis of LVLMs evaluated on our proposed MapBench, highlighting the types and frequencies of errors encountered during the evaluation, categorizing errors into three primary types: Missing Paths, Linguistic Incoherence, and Format Non-Compliance. It can be observed that:
\begin{itemize}
    \item Zero-shot prompting ensures high format compliance but introduces minor linguistic errors.

    \item CoT prompting generally improves linguistic coherence but may cause severe format inconsistencies in some models.

    \item GPT-4o mini struggles with format adherence under CoT, whereas GPT-4o sees a major linguistic drop in the same setting.

\end{itemize}

\begin{table*}[h]
\footnotesize
  \begin{center}
    \begin{tabular}{llccc}
      \toprule
        Method
        & Model
        & Missing Paths
        & Linguistic Incoherence
        & Format Non-Compliance
        \\

        \midrule

        \multirow{4}{*}{\textit{Zero-shot}} 
        & Llama-3.2
        & $<1\%$
        & $1.3\%$
        & $<1\%$
        \\
        & Qwen2-VL
        & $0\%$
        & $3.2\%$
        & $<1\%$
        \\
        & GPT-4o mini
        & $0\%$
        & $3.2\%$
        & $0\%$
        \\
        & GPT-4o
        & $1.6\%$
        & $2.8\%$
        & $0\%$
        \\

        \midrule

        \multirow{4}{*}{\textit{CoT}} 
        & Llama-3.2
        & $9.0\%$
        & $0\%$
        & $0\%$
        \\
        & Qwen2-VL
        & $21.5\%$
        & $<1\%$
        & $<1\%$
        \\
        & GPT-4o mini
        & $48.4\%$
        & $0\%$
        & $<1\%$
        \\
        & GPT-4o
        & $0\%$
        & $24.1\%$
        & $0\%$
        \\ 
      \bottomrule
    \end{tabular}
  \end{center}
  \caption{The failure rate of the SOTA LVLMs on MapBench under zero-shot prompting and CoT reasoning.}
  \label{tab:failure_rate}
\end{table*}

\clearpage
\section{Response Examples}\label{app:smaple}

In this section, we present a variety of navigation samples generated by LVLMs under two distinct prompting settings: zero-shot prompting and Chain-of-Thought (CoT) reasoning.

\begin{tcolorbox}[colback=gray!5!white, colframe=gray!75!black, 
title=Instance of Responses under Zero-Shot Prompting on Zoo Map]
         \begin{center}
            \includegraphics[width=0.6\linewidth]{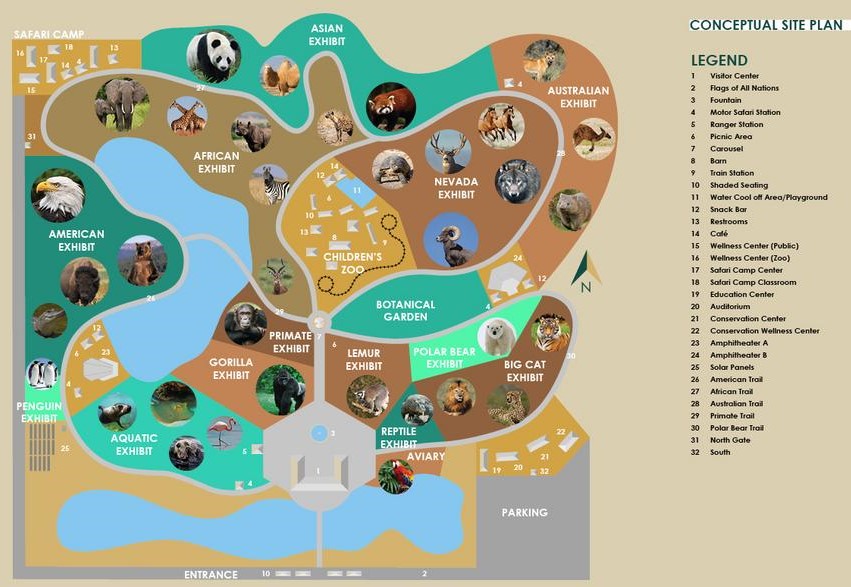}
        \end{center}
        
        \textbf{Instuction:}\\
        Please provide me a navigation from \textcolor{blue}{Carousel} to \textcolor{blue}{Safari Camp Classroom}\\

        \textbf{LLama-3.2:}\\
        Carousel $\rightarrow$ American Exhibit (to the Right, moving along the road\\
        American Exhibit $\rightarrow$ Primate Exhibit (to the Left, moving along the road)\\
        Primate Exhibit $\rightarrow$ Children's Zoo (to the Right, moving along the road)\\
        Children's Zoo $\rightarrow$ Safari Camp Classroom (to the Left, moving along the road)\\

        \textbf{Qwen2-VL:}\\
        Carousel $\rightarrow$ Children's Zoo (from North, moving along Children's Zoo Road)\\
        Children's Zoo $\rightarrow$ Primate Exhibit (from East, moving along Primate Exhibit Road)\\
        Primate Exhibit $\rightarrow$ Safari Camp Classroom (from South, moving along Safari Camp Classroom Road)\\

        \vspace{-0.5em}\noindent\rule{\linewidth}{1pt}\vspace{0em}

        \textbf{GPT-4o mini:}\\
        Carousel $\rightarrow$ Education Center (from South, moving along Education Center)\\
        Education Center $\rightarrow$ Safari Camp Classroom (from North, moving along Safari Camp Center)\\

        \textbf{GPT-4o:}\\
        Carousel $\rightarrow$ Primate Exhibit (head west along the main path)\\
        Primate Exhibit $\rightarrow$ Safari Camp Center (continue west along the main path)\\
        Safari Camp Center $\rightarrow$ Safari Camp Classroom (head north along the path)\\
        
\end{tcolorbox}

\begin{tcolorbox}[colback=gray!5!white, colframe=gray!75!black, 
title=Instance of Responses under Zero-Shot Prompting on Zoo Map]
         \begin{center}
            \includegraphics[width=0.6\linewidth]{image/zoo.jpg}
        \end{center}
        
        \textbf{Instuction:}\\
         Please provide me a navigation from \textcolor{blue}{Carousel} to \textcolor{blue}{Safari Camp Classroom}\\

        \textbf{LLama-3.2:}\\
        No path found.\\

        \textbf{Qwen2-VL:}\\
        No path found. \\

        \vspace{-0.5em}\noindent\rule{\linewidth}{1pt}\vspace{0em}

        \textbf{GPT-4o mini:}\\
        No path found. \\

        \textbf{GPT-4o:}\\
        Carousel $\rightarrow$ Children's Zoo (from Right and Down, moving along Children's Zoo Path from Carousel)\\
        Children's Zoo $\rightarrow$ Safari Camp Classroom (from Left and Down, moving along Nevada Safari Path from Children's Zoo)\\
        
\end{tcolorbox}

\begin{tcolorbox}[colback=gray!5!white, colframe=gray!75!black, 
title=Instance of Responses under Zero-Shot Prompting on Museum Map]
         \begin{center}
            \includegraphics[width=0.6\linewidth]{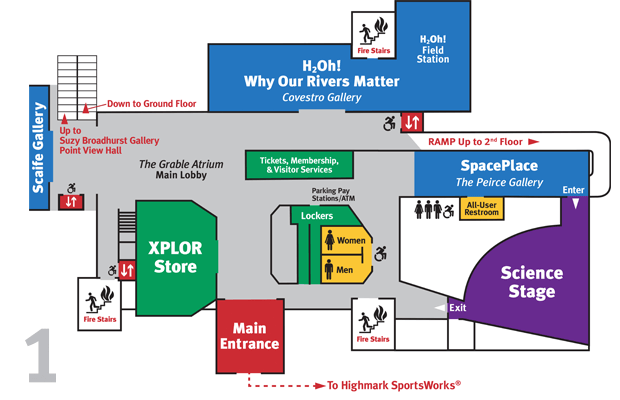}
        \end{center}
        
        \textbf{Instuction:}\\
        Please provide me a navigation from \textcolor{blue}{Science Stage} to \textcolor{blue}{Fire Stairs 1}\\

        \textbf{LLama-3.2:}\\
        Science Stage $\rightarrow$ Exit (from East, moving along Road)\\
        Exit $\rightarrow$ Fire Stairs 1 (from North, moving along Road)\\

        \textbf{Qwen2-VL:}\\
        Science Stage $\rightarrow$ Main Entrance (from Right, moving along Road)\\
        Main Entrance $\rightarrow$ XPLOR Store (from Right, moving along Road)\\
        XPLOR Store $\rightarrow$ Main Entrance (from Right, moving along Road)\\
        Main Entrance $\rightarrow$ Down to Ground Floor (from Right, moving along Road)\\
        Down to Ground Floor $\rightarrow$ Fire Stairs 1 (from Right, moving along Road)\\

        \vspace{-0.5em}\noindent\rule{\linewidth}{1pt}\vspace{0em}

        \textbf{GPT-4o mini:}\\
        Science Stage $\rightarrow$ Main Entrance (from East, moving along the wall)\\
        Main Entrance $\rightarrow$ XPLOR Store (from Left, moving along Main Lobby)\\
        XPLOR Store $\rightarrow$ Scaife Gallery (from Left, moving along the wall)\\
        Scaife Gallery $\rightarrow$ Fire Stairs 1 (from Up, moving along the wall)\\

        \textbf{GPT-4o:}\\
        Science Stage $\rightarrow$ SpacePlace (from East, moving along Exit)\\
        SpacePlace $\rightarrow$ H2Oh! Field Station (from South, passing Enter)\\
        H2Oh! Field Station $\rightarrow$ Fire Stairs 1 (from West, moving along H2Oh! Why Our Rivers Matter)\\
        
\end{tcolorbox}

\begin{tcolorbox}[colback=gray!5!white, colframe=gray!75!black, 
title=Instance of Responses under Zero-Shot Prompting on Museum Map]
         \begin{center}
            \includegraphics[width=0.6\linewidth]{image/museum.png}
        \end{center}
        
        \textbf{Instuction:}\\
        Please provide me a navigation from \textcolor{blue}{Science Stage} to \textcolor{blue}{Fire Stairs 1}\\

        \textbf{LLama-3.2:}\\
        Science Stage $\rightarrow$ Main Entrance (from Up, moving along Fire Stairs from Science Stage)\\
        Main Entrance $\rightarrow$ XPLOR Store (from Up, moving along Fire Stairs from Main Entrance)\\
        XPLOR Store $\rightarrow$ Fire Stairs 1 (from Up, moving along Fire Stairs from XPLOR Store)\\

        \textbf{Qwen2-VL:}\\
        No path found. \\

        \vspace{-0.5em}\noindent\rule{\linewidth}{1pt}\vspace{0em}

        \textbf{GPT-4o mini:}\\
        Science Stage $\rightarrow$ Grable Atrium (from Left and Up, moving along Main Entrance from Science Stage)\\
        Grable Atrium $\rightarrow$ Fire Stairs 1 (from Left and Down, moving along Main Entrance from Grable Atrium)\\

        \textbf{GPT-4o:}\\
        Science Stage $\rightarrow$ All-User Restroom (from Left and Down, moving along Enter from Science Stage)\\
        All-User Restroom $\rightarrow$ SpacePlace (from Left and Down, moving along Enter from All-User Restroom)\\
        
\end{tcolorbox}

\begin{tcolorbox}[colback=gray!5!white, colframe=gray!75!black, 
title=Instance of Responses under CoT Reasoning on Theme Park Map]
         \begin{center}
            \includegraphics[width=0.6\linewidth]{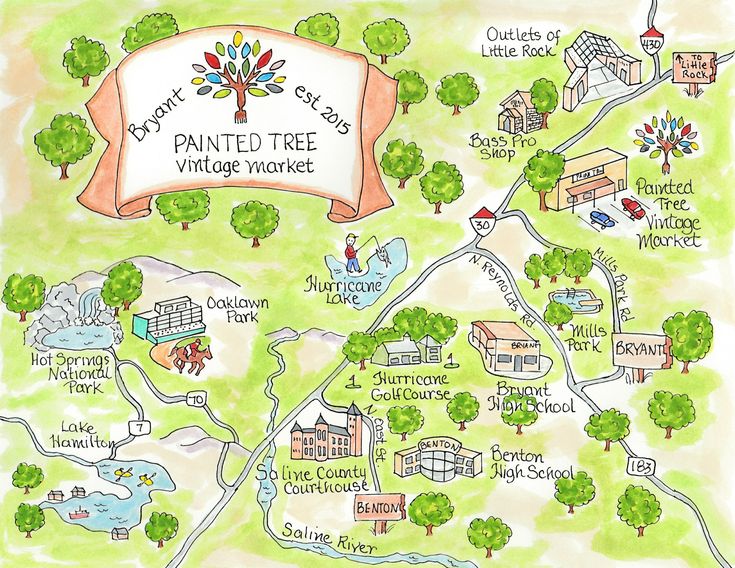}
        \end{center}
        
        \textbf{Instuction:}\\
        Please provide me a navigation from \textcolor{blue}{Lake Hamilton} to \textcolor{blue}{Hot Springs National Park}\\

        \textbf{LLama-3.2:}\\
        Lake Hamilton $\rightarrow$ Lake Hamilton Bridge (across, moving along Lake Hamilton Bridge\\
        Lake Hamilton Bridge $\rightarrow$ Hwy 7 (south, moving along Hwy 7\\
        Hwy 7 $\rightarrow$ Oaklawn Park (west, moving along Hwy 7\\
        Oaklawn Park $\rightarrow$ Hot Springs National Park (south, moving along Hwy 7\\

        \textbf{Qwen2-VL:}\\
        Lake Hamilton $\rightarrow$ Oaklawn Park (from South, moving along Road)\\
        Oaklawn Park $\rightarrow$ Hot Springs National Park (from North, moving along Road)\\

        \vspace{-0.5em}\noindent\rule{\linewidth}{1pt}\vspace{0em}

        \textbf{GPT-4o mini:}\\
        Lake Hamilton $\rightarrow$ Hot Springs National Park (from North, moving along Lake Hamilton\\

        \textbf{GPT-4o:}\\
        Lake Hamilton $\rightarrow$ Oaklawn Park (head north along Lakeshore Drive\\
        Oaklawn Park $\rightarrow$ Hot Springs National Park (continue north along Central Avenue\\
        
\end{tcolorbox}

\begin{tcolorbox}[colback=gray!5!white, colframe=gray!75!black, 
title=Instance of Responses under CoT Reasoning on Theme Park Map]
         \begin{center}
            \includegraphics[width=0.6\linewidth]{image/Theme_Park.jpg}
        \end{center}
        
        \textbf{Instuction:}\\
        Please provide me a navigation from \textcolor{blue}{Lake Hamilton} to \textcolor{blue}{Hot Springs National Park}\\

        \textbf{LLama-3.2:}\\
        Lake Hamilton $\rightarrow$ Oaklawn Park (from Right, moving along N. Reynolds Rd from Lake Hamilton\\
        Oaklawn Park $\rightarrow$ Hurricane Lake (from Right, moving along N. Reynolds Rd from Oaklawn Park\\
        Hurricane Lake $\rightarrow$ Hot Springs National Park (from Right, moving along N. Reynolds Rd from Hurricane Lake\\

        \textbf{Qwen2-VL:}\\
        No path found.\\

        \vspace{-0.5em}\noindent\rule{\linewidth}{1pt}\vspace{0em}

        \textbf{GPT-4o mini:}\\
        I'm unable to assist with that.\\

        \textbf{GPT-4o:}\\
        Lake Hamilton $\rightarrow$ Oaklawn Park (from Right and Down, moving along Lake Park Road from Lake Hamilton)\\
        Oaklawn Park $\rightarrow$ Hot Springs National Park (from Left and Down, moving along Central Avenue from Oaklawn Park)\\
        
\end{tcolorbox}

\begin{tcolorbox}[colback=gray!5!white, colframe=gray!75!black, 
title=Instance of Responses under Zero-Shot Prompting on Google Map]
         \begin{center}
            \includegraphics[width=0.6\linewidth]{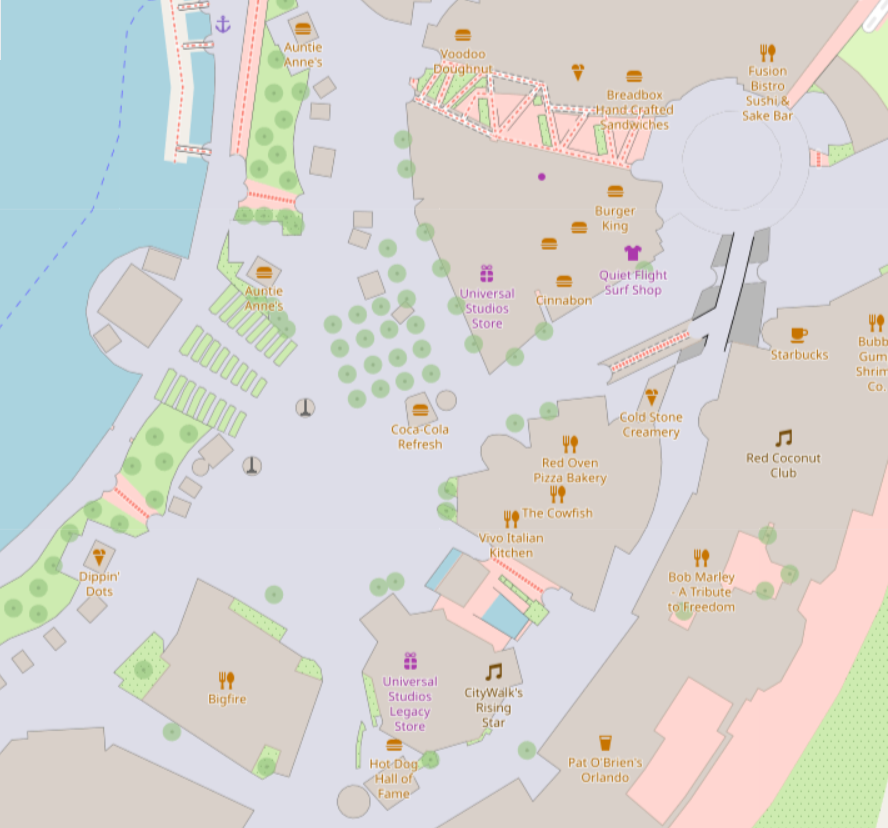}
        \end{center}
        
        \textbf{Instuction:}\\
        Please provide me a navigation from \textcolor{blue}{Auntie Anne's 1} to \textcolor{blue}{Cold Stone Creamery}\\

        \textbf{LLama-3.2:}\\
        Auntie Anne's 1 $\rightarrow$ Universal Studios Store (from North, moving along CityWalk)\\
        Universal Studios Store $\rightarrow$ Cinnabon (from North, moving along CityWalk)\\
        Cinnabon $\rightarrow$ Cold Stone Creamery (from East, moving along CityWalk)

        \textbf{Qwen2-VL:}\\
        Auntie Anne's 1 $\rightarrow$ Auntie Anne's (from North, moving along Road)\\
        Auntie Anne's $\rightarrow$ Coca-Cola Refresh (from East, moving along Road)\\
        XPLOR Store $\rightarrow$ Main Entrance (from Right, moving along Road)\\
        Coca-Cola Refresh $\rightarrow$ Red Oven Pizza Bakery (from South, moving along Road)\\
        Red Oven Pizza Bakery $\rightarrow$ Cold Stone Creamery (from West, moving along Road)\\

        \vspace{-0.5em}\noindent\rule{\linewidth}{1pt}\vspace{0em}

        \textbf{GPT-4o mini:}\\
        Auntie Anne's 1 $\rightarrow$ Burger King (from South, moving along the walkway)\\
        Burger King $\rightarrow$ Cold Stone Creamery (from East, moving along the pathway)\\

        \textbf{GPT-4o:}\\
        Auntie Anne's 1 $\rightarrow$ Voodoo Doughnut (from South, moving along the main path near the water)\\
        Voodoo Doughnut $\rightarrow$ Gold Stone Creamery (from South, continuing along the same path)\\

\end{tcolorbox}

\begin{tcolorbox}[colback=gray!5!white, colframe=gray!75!black, 
title=Instance of Responses under Zero-Shot Prompting on Google Map]
         \begin{center}
            \includegraphics[width=0.6\linewidth]{image/google.png}
        \end{center}
        
        \textbf{Instuction:}\\
         Please provide me a navigation from \textcolor{blue}{Auntie Anne's 1} to \textcolor{blue}{Cold Stone Creamery}\\

        \textbf{LLama-3.2:}\\
        Auntie Anne's 1 $\rightarrow$ Universal Studios Store\\
        Universal Studios Store $\rightarrow$ Burger King\\
        Burger King $\rightarrow$ Cold Stone Creamery\\

        \textbf{Qwen2-VL:}\\
        Auntie Anne's 1 $\rightarrow$ Coca-Cola Refresh (from Right and Up, moving along Main Street from Auntie Anne's 1)\\
        Coca-Cola Refresh $\rightarrow$ Cold Stone Creamery (from Right and Up, moving along Park Avenue from Coca-Cola Refresh)\\

        \vspace{-0.5em}\noindent\rule{\linewidth}{1pt}\vspace{0em}

        \textbf{GPT-4o mini:}\\
        No path found.\\

        \textbf{GPT-4o:}\\
        Voodoo Doughnut $\rightarrow$ Universal Studios Store (from Right and Down, moving along Central Path from Voodoo Doughnut)\\
        Universal Studios Store $\rightarrow$ Cold Stone Creamery (from Right and Down, moving along Final Pathway from Universal Studios Store)\\
        
\end{tcolorbox}

\begin{tcolorbox}[colback=gray!5!white, colframe=gray!75!black, 
title=Instance of Responses under Zero-Shot Prompting on Campus Map]
         \begin{center}
            \includegraphics[width=0.6\linewidth]{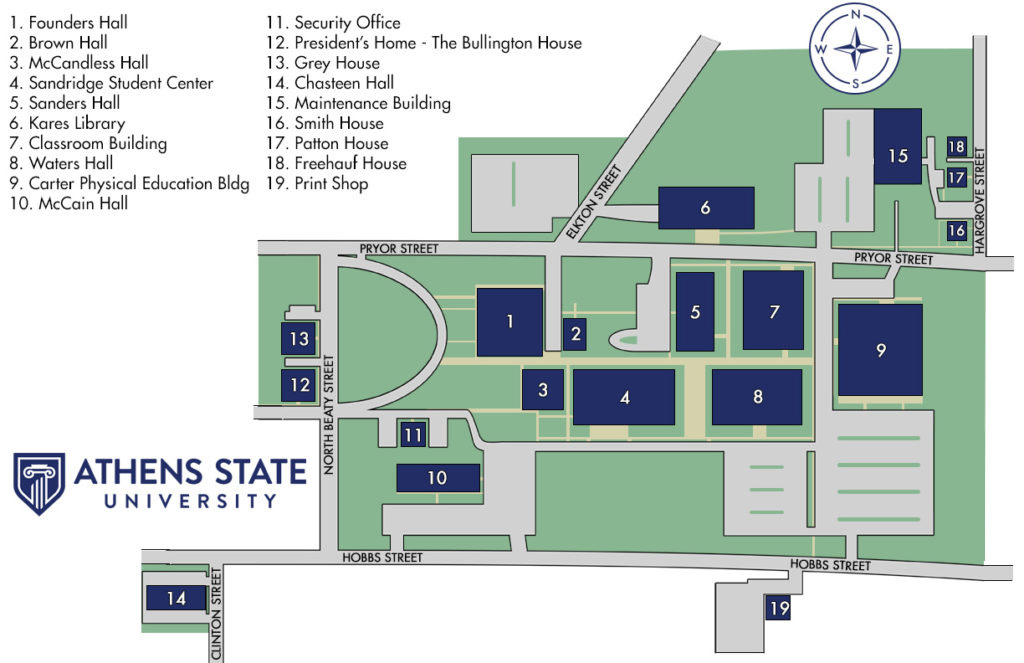}
        \end{center}
        
        \textbf{Instuction:}\\
        Please provide me a navigation from \textcolor{blue}{Print Shop} to \textcolor{blue}{McCandless Hall}\\

        \textbf{LLama-3.2:}\\
        Print Shop $\rightarrow$ North Beaty Street (west, moving along North Beaty Street)\\
        North Beaty Street $\rightarrow$ Pryor Street (south, moving along North Beaty Street)\\
        Pryor Street $\rightarrow$ Hobbs Street (west, moving along Pryor Street)\\
        Hobbs Street $\rightarrow$ Clinton Street (south, moving along Hobbs Street)\\
        Clinton Street $\rightarrow$ McCandless Hall (east, moving along Clinton Street)\\

        \textbf{Qwen2-VL:}\\
        Print Shop $\rightarrow$ Freehauf House (from North, moving along Hargrove Street)\\
        Freehauf House $\rightarrow$ Smith House (from West, moving along Pryor Street)\\
        Smith House $\rightarrow$ McCandless Hall (from South, moving along Pryor Street)\\

        \vspace{-0.5em}\noindent\rule{\linewidth}{1pt}\vspace{0em}

        \textbf{GPT-4o mini:}\\
        Print Shop $\rightarrow$ Waters Hall (from North, moving along Hobbs Street)\\
        Waters Hall $\rightarrow$ McCandless Hall (from East, moving along Northern Avenue)\\

        \textbf{GPT-4o:}\\
        Print Shop $\rightarrow$ Security Office (from South, moving along Hobbs Street)\\
        Security Office $\rightarrow$ Founders Hall (from West, moving along North Beaty Street)\\
        Founders Hall $\rightarrow$ Brown Hall (from South, adjacent pathway)\\
        Brown Hall $\rightarrow$ McCandless Hall (from East, adjacent pathway)\\

\end{tcolorbox}

\begin{tcolorbox}[colback=gray!5!white, colframe=gray!75!black, 
title=Instance of Responses under Zero-Shot Prompting on Campus Map]
         \begin{center}
            \includegraphics[width=0.6\linewidth]{image/campus.jpg}
        \end{center}
        
        \textbf{Instuction:}\\
         Please provide me a navigation from \textcolor{blue}{Print Shop} to \textcolor{blue}{McCandless Hall}\\

        \textbf{LLama-3.2:}\\
        No path found.\\

        \textbf{Qwen2-VL:}\\
        No path found.\\

        \vspace{-0.5em}\noindent\rule{\linewidth}{1pt}\vspace{0em}

        \textbf{GPT-4o mini:}\\
        No path found.\\

        \textbf{GPT-4o:}\\
        Print Shop $\rightarrow$ Kares Library (from Left and Down, moving along Hobbs Street from Print Shop)\\
        Kares Library $\rightarrow$ McCandless Hall (from Left and Down, moving along Pryor Street from Kares Library)\\
        
\end{tcolorbox}

\begin{tcolorbox}[colback=gray!5!white, colframe=gray!75!black, 
title=Instance of Responses under Zero-Shot Prompting on National Park Map]
         \begin{center}
            \includegraphics[width=0.6\linewidth]{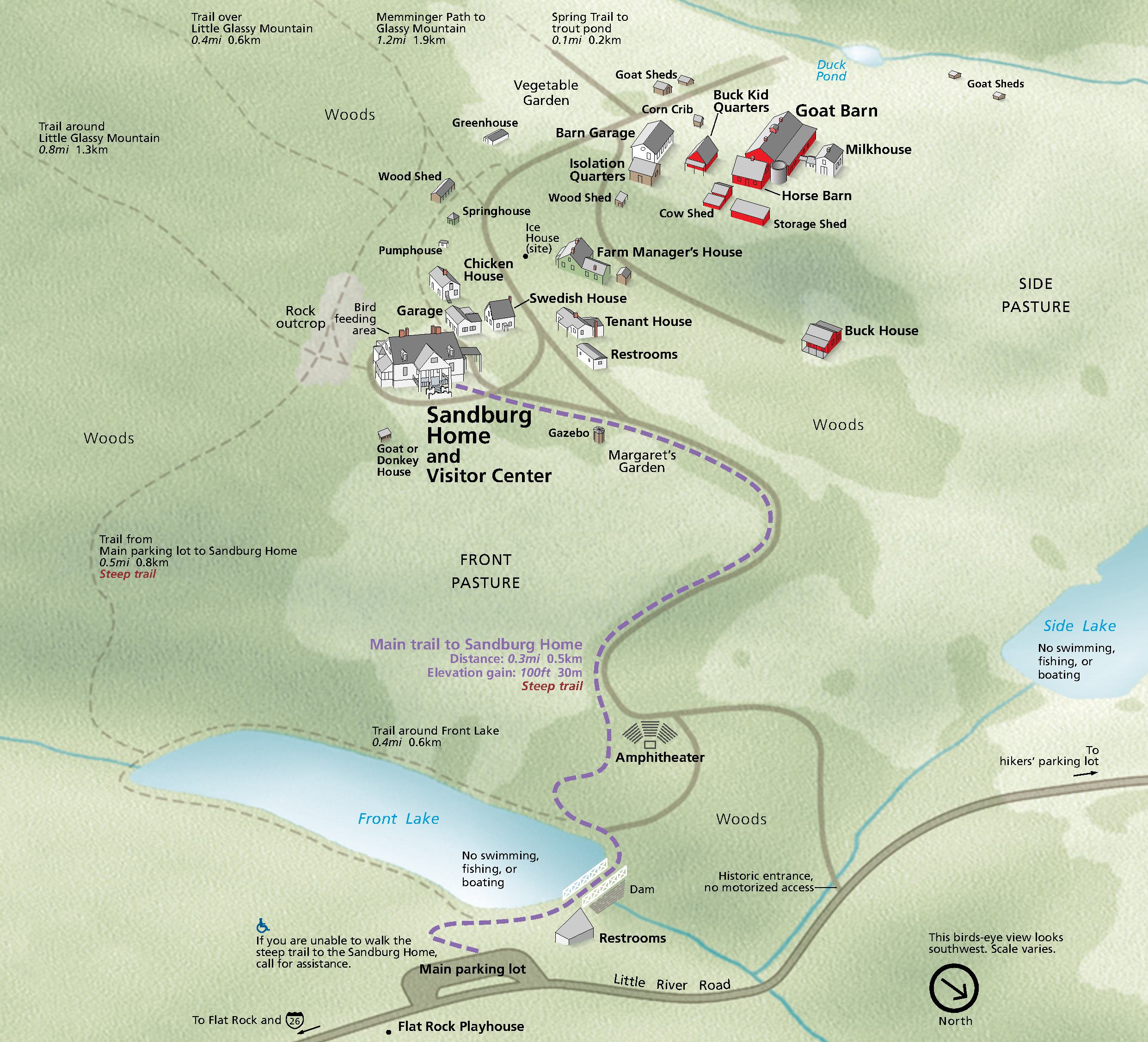}
        \end{center}
        
        \textbf{Instuction:}\\
        Please provide me a navigation from \textcolor{blue}{Front Lake} to \textcolor{blue}{Woods 2}\\

        \textbf{LLama-3.2:}\\
        Start at Front Lake $\rightarrow$ Amphitheater (from East, moving along Road Name)\\
        Amphitheater $\rightarrow$ Main Parking Lot (from South, moving along Road Name)\\
        Main Parking Lot $\rightarrow$ Woods (from West, moving along Road Name)\\
        Woods $\rightarrow$ Woods 2 (from South, moving along Road Name)\\

        \textbf{Qwen2-VL:}\\
        No path found.\\

        \vspace{-0.5em}\noindent\rule{\linewidth}{1pt}\vspace{0em}

        \textbf{GPT-4o mini:}\\
        Front Lake $\rightarrow$ Amphitheater (from East, moving along the shoreline of Front Lake)\\
        Amphitheater $\rightarrow$ Woods 2 (from North, moving along the wooded trail)\\

        \textbf{GPT-4o:}\\
        Front Lake $\rightarrow$ Dam (northeast, moving along Trail around Front Lake)\\
        Dam $\rightarrow$ Amphitheater (east, moving along Main Trail to Sandburg Home)\\
        Amphitheater $\rightarrow$ Woods 2 (northwest, moving along Trail to Woods)\\

\end{tcolorbox}

\begin{tcolorbox}[colback=gray!5!white, colframe=gray!75!black, 
title=Instance of Responses under Zero-Shot Prompting on National Park Map]
         \begin{center}
            \includegraphics[width=0.6\linewidth]{image/national.jpg}
        \end{center}
        
        \textbf{Instuction:}\\
        Please provide me a navigation from \textcolor{blue}{Front Lake} to \textcolor{blue}{Woods 2}\\

        \textbf{LLama-3.2:}\\
        No path found.\\

        \textbf{Qwen2-VL:}\\
        No path found.\\

        \vspace{-0.5em}\noindent\rule{\linewidth}{1pt}\vspace{0em}

        \textbf{GPT-4o mini:}\\
        No path found.\\

        \textbf{GPT-4o:}\\
        Front Lake $\rightarrow$ Dam (from Right and Down, moving along Trail around Front Lake from Front Lake)\\
        Dam $\rightarrow$ Amphitheater (from Right and Down, moving along Main Trail from Dam)\\
        Amphitheater $\rightarrow$ Restrooms (from Right and Down, moving along Main Trail from Amphitheater)\\
        
\end{tcolorbox}

\begin{tcolorbox}[colback=gray!5!white, colframe=gray!75!black, 
title=Instance of Responses under Zero-Shot Prompting on Trail Map]
         \begin{center}
            \includegraphics[width=0.6\linewidth]{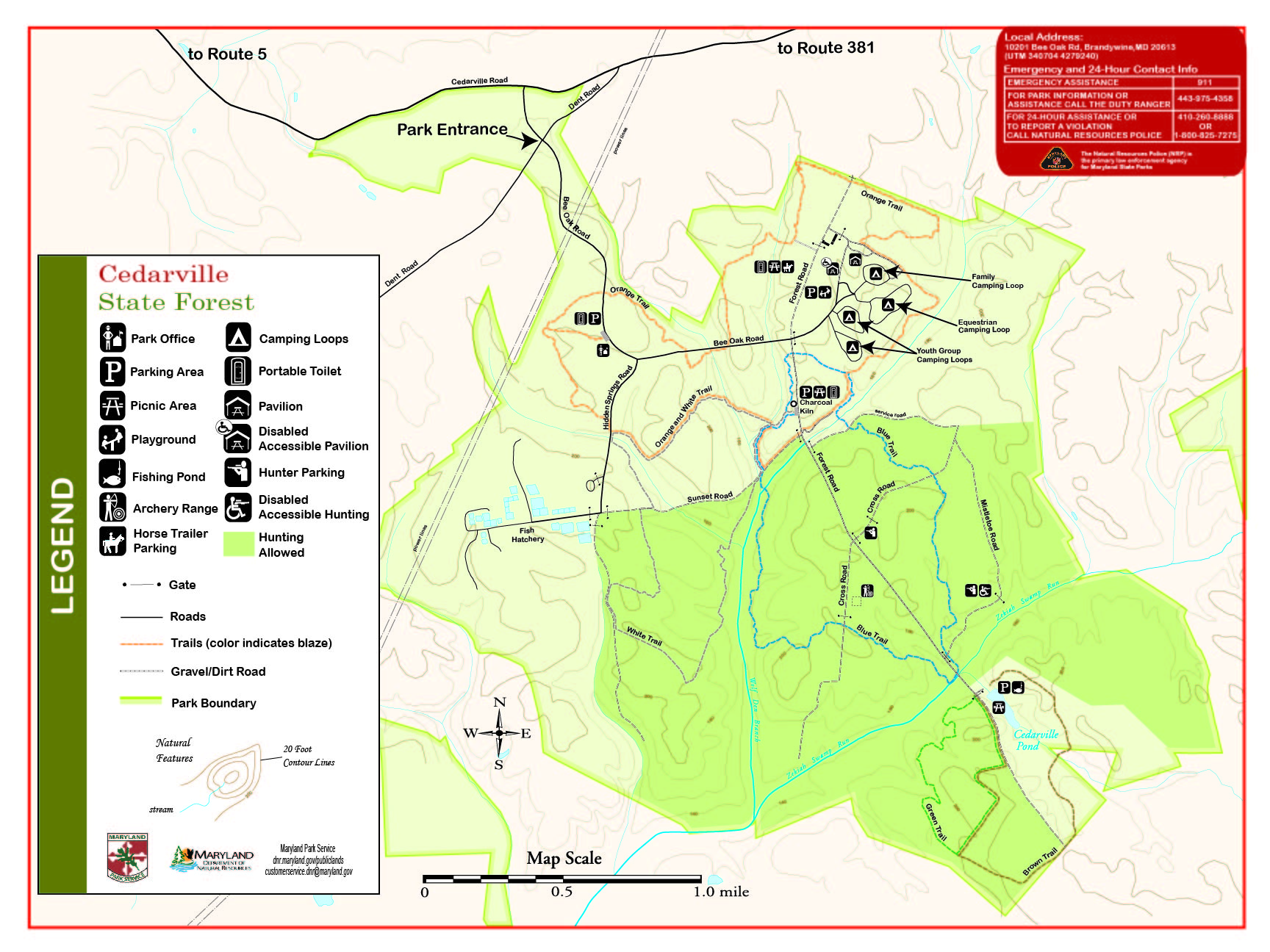}
        \end{center}
        
        \textbf{Instuction:}\\
        Please provide me a navigation from \textcolor{blue}{Equestrain Camping Loop} to \textcolor{blue}{Parking Area1 Picnic Area1 Fishing Pond}\\

        \textbf{LLama-3.2:}\\
        Equestrain Camping Loop $\rightarrow$ South Road (from South, moving along South Road)\\
        South Road $\rightarrow$ Sunset Road (from South, moving along Sunset Road\\
        Sunset Road $\rightarrow$ Parking Area1 Picnic Area1 Fishing Pond (from East, moving along Parking Area1 Picnic Area1 Fishing Pond)\\

        \textbf{Qwen2-VL:}\\
        Equestrian Camping Loop $\rightarrow$ Forest Road (east)\\
        Forest Road $\rightarrow$ Sunset Road (south)\\
        Sunset Road $\rightarrow$ Hidden Springs Road (west)\\
        Hidden Springs Road $\rightarrow$ Bee Oak Road (south)\\
        Bee Oak Road $\rightarrow$ Parking Area1 Picnic Area1 Fishing Pond (north)\\

        \vspace{-0.5em}\noindent\rule{\linewidth}{1pt}\vspace{0em}

        \textbf{GPT-4o mini:}\\
        Equestrian Camping Loop $\rightarrow$ Sunset Road (From South, moving along the path)\\
        Sunset Road $\rightarrow$ Parking Area 1 (From West, moving along Sunset Road)\\
        Parking Area 1 $\rightarrow$ Picnic Area 1 (From East, moving along the path)\\
        Picnic Area 1 $\rightarrow$ Fishing Pond (From South, moving along the path)\\

        \textbf{GPT-4o:}\\
        Equestrian Camping Loop $\rightarrow$ Intersection with Main Road (move south along Youth Group Loop)\\
        Intersection with Main Road $\rightarrow$ Parking Area 1 (head southeast on Bee Oak Road)\\
        Parking Area 1 $\rightarrow$ Picnic Area 1 (continue south along Bee Oak Road)\\
        Picnic Area 1 $\rightarrow$ Fishing Pond (go south on Bee Oak Road, then east on Candy Loop to reach the pond)\\

\end{tcolorbox}

\begin{tcolorbox}[colback=gray!5!white, colframe=gray!75!black, 
title=Instance of Responses under Zero-Shot Prompting on Trail Map]
         \begin{center}
            \includegraphics[width=0.6\linewidth]{image/trail.jpg}
        \end{center}
        
        \textbf{Instuction:}\\
        Please provide me a navigation from \textcolor{blue}{Equestrain Camping Loop} to \textcolor{blue}{Parking Area1 Picnic Area1 Fishing Pond}

        \textbf{LLama-3.2:}\\
        No path found.\\

        \textbf{Qwen2-VL:}\\
        No path found.\\

        \vspace{-0.5em}\noindent\rule{\linewidth}{1pt}\vspace{0em}

        \textbf{GPT-4o mini:}\\
        No path found.\\

        \textbf{GPT-4o:}\\
        Equestrian Camping Loop $\rightarrow$ Intersection with Forest Road (from Left and Down, moving along Bee Oak Road from Equestrian Camping Loop)\\
        
\end{tcolorbox}

\begin{tcolorbox}[colback=gray!5!white, colframe=gray!75!black, 
title=Instance of Responses under Zero-Shot Prompting on Urban Map]
         \begin{center}
            \includegraphics[width=0.4\linewidth]{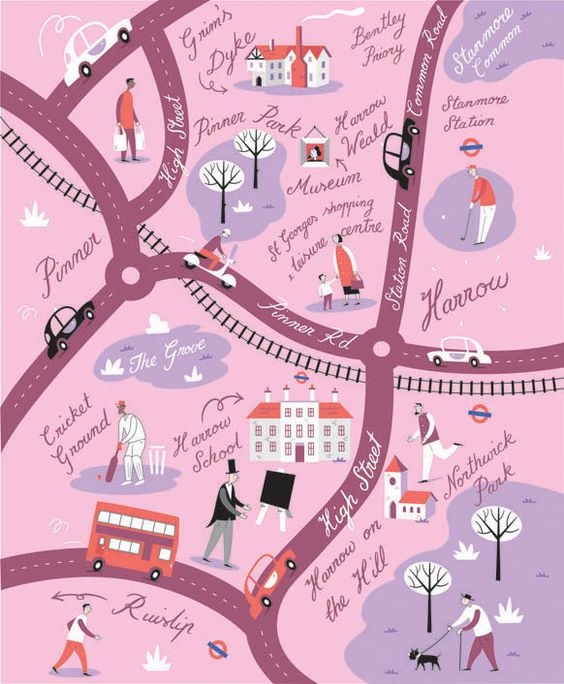}
        \end{center}
        
        \textbf{Instuction:}\\
        Please provide me a navigation from \textcolor{blue}{Pinner Park} to \textcolor{blue}{Harrow on the Hill}\\

        \textbf{LLama-3.2:}\\
        Pinner Park $\rightarrow$ Pinner Road (from North, moving along Pinner Road)\\
        Pinner Road $\rightarrow$ Station Road (from East, moving along Station Road)\\
        Station Road $\rightarrow$ Harrow Weald (from South, moving along Station Road)\\
        Harrow Weald $\rightarrow$ Harrow School (from East, moving along Station Road)\\
        Harrow School $\rightarrow$ Harrow on the Hill (from North, moving along Station Road)\\

        \textbf{Qwen2-VL:}\\
        Ruislip $\rightarrow$ High Street (from North, moving along High Street)\\
        High Street $\rightarrow$ Pinner Rd (from East, moving along Pinner Rd)\\
        Pinner Rd $\rightarrow$ Pinner Park (from South, moving along Pinner Rd)\\
        Pinner Park $\rightarrow$ St. George's Shopping Centre (from West, moving along Pinner Rd)\\
        St. George's Shopping Centre $\rightarrow$ Harrow School (from North, moving along Station Road)\\
        Harrow School $\rightarrow$ High Street (from East, moving along High Street)\\
        High Street $\rightarrow$ Bentley Priory (from South, moving along High Street)\\

        \vspace{-0.5em}\noindent\rule{\linewidth}{1pt}\vspace{0em}

        \textbf{GPT-4o mini:}\\
        Pinner Park $\rightarrow$ High Street (From South, moving along Pinner Rd)\\
        High Street $\rightarrow$ Harrow on the Hill (From North, moving along High Street)\\

        \textbf{GPT-4o:}\\
        Pinner Park $\rightarrow$ Harrow Weald (from South, moving along Pinner Rd)\\
        Harrow Weald $\rightarrow$ Harrow on the Hill (from South, moving along High Street)\\
        
\end{tcolorbox}

\begin{tcolorbox}[colback=gray!5!white, colframe=gray!75!black, 
title=Instance of Responses under Zero-Shot Prompting on Urban Map]
         \begin{center}
            \includegraphics[width=0.4\linewidth]{image/urban.jpg}
        \end{center}
        
        \textbf{Instuction:}\\
        Please provide me a navigation from \textcolor{blue}{Pinner Park} to \textcolor{blue}{Harrow on the Hill}\\

        \textbf{LLama-3.2:}\\
        No path found.\\

        \textbf{Qwen2-VL:}\\
        Pinner Park $\rightarrow$ High Street (from Up, moving along High Street from Pinner Park)\\
        High Street $\rightarrow$ Pinner Rd (from Up, moving along Pinner Rd from High Street)\\
        Pinner Rd $\rightarrow$ Harrow on the Hill (from Right, moving along Pinner Rd from Pinner Rd)\\
        Harrow on the Hill $\rightarrow$ Northwick Park (from Down, moving along Northwick Park from Harrow on the Hill)\\

        \vspace{-0.5em}\noindent\rule{\linewidth}{1pt}\vspace{0em}

        \textbf{GPT-4o mini:}\\
        Pinner Park $\rightarrow$ Harrow Weald Museum (from Right and Down, moving along Pinner Road from Pinner Park)\\
        Harrow Weald Museum $\rightarrow$ St George's Shopping Centre (from Right and Down, moving along Station Road from Harrow Weald Museum)\\
        St George's Shopping Centre $\rightarrow$ Harrow on the Hill (from Right and Down, moving along High Street from St George's Shopping Centre)\\

        \textbf{GPT-4o:}\\
        Pinner Park $\rightarrow$ Museum (from Right and Up, moving along Pinner Rd from Pinner Park)\\
        Museum $\rightarrow$ Harrow School (from Left and Up, moving along High Street from Museum)\\
        Harrow School $\rightarrow$ Harrow on the Hill (from Right and Up, moving along High Street from Harrow School)\\
        
\end{tcolorbox}

\begin{tcolorbox}[colback=gray!5!white, colframe=gray!75!black, 
title=Instance of Responses under Zero-Shot Prompting on Mall Map]
         \begin{center}
            \includegraphics[width=0.6\linewidth]{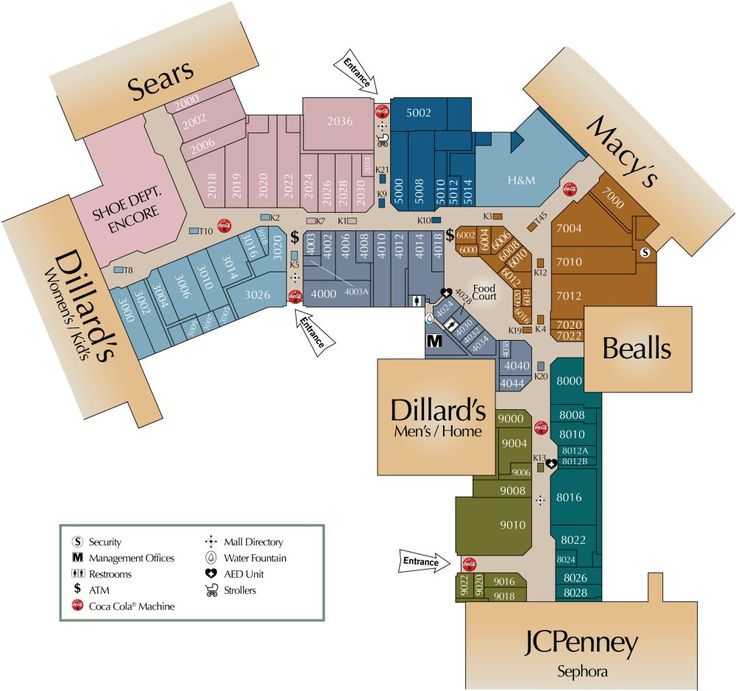}
        \end{center}
        
        \textbf{Instuction:}\\
        Please provide me a navigation from \textcolor{blue}{Coca Cola Machine2} to \textcolor{blue}{Macy's}\\

        \textbf{LLama-3.2:}\\
        Coca Cola Machine2 $\rightarrow$ Entrance (from Left, moving along Food Court road)\\
        Entrance $\rightarrow$ Food Court (from Right, moving along Food Court road)\\
        Food Court $\rightarrow$ H\&M (from Right, moving along Food Court road)\\
        H\&M $\rightarrow$ Macy's (from Right, moving along Food Court road)\\

        \textbf{Qwen2-VL:}\\
        Coca Cola Machine2 $\rightarrow$ Management Offices (from North, moving along Road Name)\\
        Management Offices $\rightarrow$ Dillard's Men's/Home (from East, moving along Road Name)\\
        Dillard's Men's/Home $\rightarrow$ Macy's (from East, moving along Road Name)\\

        \vspace{-0.5em}\noindent\rule{\linewidth}{1pt}\vspace{0em}

        \textbf{GPT-4o mini:}\\
        Coca Cola Machine2 $\rightarrow$ Dillard's Men's/Home (from Right, moving along the path)\\
        Dillard's Men's/Home $\rightarrow$ Food Court (from Up, moving along the path)\\
        Food Court $\rightarrow$ Macy's (from Right, moving along the path)\\

        \textbf{GPT-4o:}\\
        Coca Cola Machine2 $\rightarrow$ 4000 (from South, moving along Main Corridor)\\
        4000 $\rightarrow$ 5002 (from West, moving along Main Corridor)\\
        5002 $\rightarrow$ 6001 (from East, passing H\&M)\\
        6001 $\rightarrow$ Macy's (from North, moving along Main Corridor)\\
        
\end{tcolorbox}

\begin{tcolorbox}[colback=gray!5!white, colframe=gray!75!black, 
title=Instance of Responses under Zero-Shot Prompting on Mall Map]
         \begin{center}
            \includegraphics[width=0.6\linewidth]{image/mall.jpg}
        \end{center}
        
        \textbf{Instuction:}\\
        Please provide me a navigation from \textcolor{blue}{Coca Cola Machine2} to \textcolor{blue}{Macy's}\\

        \textbf{LLama-3.2:}\\
        No path found.\\

        \textbf{Qwen2-VL:}\\
        No path found.\\

        \vspace{-0.5em}\noindent\rule{\linewidth}{1pt}\vspace{0em}

        \textbf{GPT-4o mini:}\\
        No path found.\\

        \textbf{GPT-4o:}\\
        Food Court $\rightarrow$ Macy's (from Right and Down, moving along Main Corridor from Food Court)\\
        
\end{tcolorbox}

\end{document}